\documentclass{article}

\usepackage[english]{babel}
\usepackage{colortbl}
\usepackage[utf8]{inputenc}
\usepackage{comment}
\usepackage{geometry}
\usepackage{lipsum}
\usepackage{fancyhdr}
\usepackage{graphicx}
\usepackage{xcolor}
\usepackage{hyperref}
\usepackage{amsmath}
\usepackage{amsthm}
\usepackage{amsfonts}
\usepackage{subcaption} 
\usepackage{nomencl}
\usepackage{etoolbox}
\usepackage{footmisc}
\newtheorem{proposition}{Proposition}[section]
\usepackage[numbers,sort&compress]{natbib}
\usepackage{float}
\usepackage{enumitem}
\usepackage{tikz}
\usepackage[toc,page]{appendix}
\usepackage{longtable}
\usepackage{blkarray}
\usepackage{booktabs}
\newtheorem{theorem}{Theorem}[section]
\newtheorem{corollary}{Corollary}[theorem]

\usepackage{tabularx}
\usetikzlibrary{datavisualization}
\usetikzlibrary{datavisualization.formats.functions}
\usepackage{mathtools}

\DeclarePairedDelimiter\floor{\lfloor}{\rfloor}

\definecolor{blueLAMPS}{rgb}{0.215, 0.376, 0.572}
\definecolor{orangeLAMPS}{rgb}{0.886, 0.462, 0}

\renewcommand\nomgroup[1]{%
	\item[\bfseries
	\ifstrequal{#1}{T}{Séries temporais}{%
		\ifstrequal{#1}{S}{Conjuntos}{%
			\ifstrequal{#1}{F}{Funções}{%
				\ifstrequal{#1}{O}{Outros símbolos}{}}}}%
	]}
\usepackage[utf8]{inputenc}
\usepackage{listings}
\usepackage{xcolor}
\usepackage{algorithm} 
\usepackage{algpseudocode} 
\usepackage{amsmath}
\usepackage{textcomp}

\definecolor{codegreen}{rgb}{0,0.6,0}
\definecolor{codegray}{rgb}{0.5,0.5,0.5}
\definecolor{codepurple}{rgb}{0.58,0,0.82}
\definecolor{backcolour}{rgb}{0.95,0.95,0.92}

\lstdefinestyle{mystyle}{
    backgroundcolor=\color{backcolour},   
    commentstyle=\color{codegreen},
    keywordstyle=\color{magenta},
    numberstyle=\tiny\color{codegray},
    stringstyle=\color{codepurple},
    basicstyle=\ttfamily\footnotesize,
    breakatwhitespace=false,         
    breaklines=true,                 
    captionpos=b,                    
    keepspaces=true,                 
    numbers=left,                    
    numbersep=5pt,                  
    showspaces=false,                
    showstringspaces=false,
    showtabs=false,                  
    tabsize=2
}
\newcolumntype{C}[1]{>{\centering\let\newline\\\arraybackslash\hspace{0pt}}m{#1}}
\title{Time Series Analysis by State Space Learning}
\author{André Ramos, Davi Valladão, Alexandre Street}

\begin{document}

\maketitle

\begin{abstract}
Time series analysis by state-space models is widely used in forecasting and extracting unobservable components like level, slope, and seasonality, along with explanatory variables. However, their reliance on traditional Kalman filtering frequently hampers their effectiveness, primarily due to Gaussian assumptions and the absence of efficient subset selection methods to accommodate the multitude of potential explanatory variables in today's big-data applications. Our research introduces the State Space Learning (SSL), a novel framework and paradigm that leverages the capabilities of statistical learning to construct a comprehensive framework for time series modeling and forecasting. 
By utilizing a regularized high-dimensional regression framework, our approach jointly extracts typical time series unobservable components, detects and addresses outliers, and selects the influence of exogenous variables within a high-dimensional space in polynomial time and global optimality guarantees. Through a controlled numerical experiment, we demonstrate the superiority of our approach in terms of subset selection of explanatory variables accuracy compared to relevant benchmarks. We also present an intuitive forecasting scheme and showcase superior performances relative to traditional time series models using a dataset of 48,000 monthly time series from the M4 competition.
We extend the applicability of our approach to reformulate any linear state space formulation featuring time-varying coefficients into high-dimensional regularized regressions, expanding the impact of our research to other engineering applications beyond time series analysis. Finally, our proposed methodology is implemented within the Julia open-source package, ``StateSpaceLearning.jl". 
\end{abstract}


\section{Introduction}\label{Introduction}

Time series analysis by state space models is a flexible framework for capturing non-observable components such as level, trend, and seasonality, as well as exogenous features. This makes them an attractive choice for modeling temporal data across various domains, including energy, control, telecommunications, signal processing, and business, just to mention a few. According to \citep{Durbin_Koopman}, the state space framework confers enhanced flexibility compared to conventional models, exemplified by ARIMA \citep{boxjen76}). This heightened flexibility arises from its capacity to circumvent certain assumptions, such as the imposition of stationarity. 

State space modeling, coupled with the Kalman filter technique \cite{Kalman}, has emerged as the standard approach for estimating models with unobserved components with or without exogenous variables. Nevertheless, this method has several limitations that restrict its applicability in real-world scenarios. First, the existing methodology relies on non-linear optimization techniques, which may provide estimates that are locally optimal. However, the quality of these estimates is heavily dependent on the user's initial guesses, which may not always be well-informed. In practice, users may resort to trial-and-error or stochastic generation procedures, which can lead to suboptimal and computationally expensive results. Second, despite the state space model's ability to incorporate explanatory variables, it lacks the functionality to conduct built-in subset selection, which can limit its performance in high-dimensional settings characterized by numerous potential explanatory variables (a common feature in today's data-rich environments). Third, although the state space framework is able to detect outliers by studying the auxiliary residuals, as explained in \cite{Durbin_Koopman}, it is done in a heuristic two-step process, which may lead to loss of information.

As such, it is of utmost importance for the time series community to develop novel and effective approaches that can mitigate the limitations of the current methodologies while preserving key features from traditional models. 
Therefore, the objective of this work is to propose a novel, efficient, and integrated framework for time series in state-space format, named State Space Learning (SSL). Based on this new formulation, a new paradigm is unlocked for non-stationary time-series modeling. The main contributions of the proposed framework concerning the existing technical literature on the subject are the following:

\begin{itemize}
  \item a novel formulation for structural models (with unobserved components of level, slope, and seasonality) based on a high-dimensional regularized regression on shifted steps, ReLus, and alternating functions; 
  \item a computationally efficient (in polynomial time) estimation method for state-space models based on convex optimization that ensures global optimality;
  \item an effective and integrated method to detect and handle outliers within the proposed SSL estimation process;
  \item an optimal subset selection methodology applicable to non-stationary time series;
  \item a k-step ahead out-of-sample forecasting approach (extrapolation) with improved forecasting capacity compared to traditional state space benchmarks;
  \item a new interpolation method to handle missing values in non-stationary time series;
  \item effective way to fully initialize basic Gaussian structural models with high-quality innovations, initial conditions, hyperparameters, and selected explanatory variables;
  \item a generalization that can be applied to any linear state-space model with time-varying coefficients;
  \item an open-source package of the proposed method in Julia:``StateSpaceLearning.jl".
\end{itemize}

We conduct an extensive set of numerical experiments using synthetic and real-world time series that corroborate, with substantial empirical evidence, the robustness, predictive power, and model selection capabilities of the proposed framework. Based on our results, we believe the proposed approach represents a significant advancement over the existing methods and has the potential to substantially enhance the accuracy and reliability of temporal data modeling in diverse fields.

\subsection{Illustrative example}

The basic idea behind our approach is to establish that state space models can be reformulated as high-dimensional regression models by dismounting the model recursion and be effectively estimated using regularization techniques such as Elastic Net \cite{elasticnet}. 

To illustrate this point, we begin by introducing a simple local level model

\begin{align}
     y_t &= \mu_t + \varepsilon_t,  \label{ILE1}\\
     \mu_{t} &= \mu_{t-1} + \eta_{t} ,\label{ILE2}
 \end{align} where $y_t$ is the dependent variable being estimated, $\mu_t$ is the stochastic level component, $\varepsilon_t$ is the observation error and $\eta_t$ is the level component noise. 

 Now, we reformulate the model (\ref{ILE1}--\ref{ILE2}) by iterating the level component over time as 

\begin{align}
    y_t = \mu_1 + \sum_{\tau=2}^{t}\eta_\tau + \varepsilon_t,\label{ILE3}
 \end{align}

Note that Equation \ref{ILE3} can be written in a vectorized form 
\begin{align}
    {\bf y} = \mu_1 + {\bf X}\eta + \varepsilon,\label{LLM_ref_vec}
\end{align} where ${\bf y} = (y_1,\ldots,y_T)$, ${\eta} = (\eta_2,\ldots,\eta_T)$ and
\begin{align*}
{\bf X}=\begin{bmatrix}
0 & 0 & \cdots & 0\\
1 & 0 & \cdots & 0\\
1 & 1 & \cdots & 0\\
\cdots & \cdots & \cdots & \cdots \\
1 & 1 & \cdots & 0 \\
1 & 1 & \cdots & 1 \\
\end{bmatrix},
\end{align*} with each column is referent to one $\eta_\tau$ coefficient. 
Note that Equation \ref{LLM_ref_vec} can be interpreted as a high dimension regression and each regressor is a step function that starts at time $\tau = 1,\ldots,T$. 
To handle the large number of features in the problem, we estimate a regularized model using the following regularized regression formulation
 \begin{align}
     \underset{\mu_1,\eta}{\text{minimize }} & \bigl\|{\bf y} -\mu_1 - {\bf X}\eta \bigr\|_2^2 + \lambda\Bigl[(1-\alpha)\|\eta\|_2^2\,/\,2+\alpha\|\eta\|_1\Bigr], 
 \end{align} 
 where $ \lambda > 0$ and $ 0 \leq \alpha \leq 1$.

As delineated in the Introduction, it is noteworthy that the SSL formulation of the local-level model exhibits an underlying high-dimensional regression form that comprises a large collection of step functions as potential regressors. 
This idea is extended to a more general class of structural models with unobserved components, such as slope, level, seasonality, outliers (as we demonstrate in Section \ref{meth}), and exogenous explanatory variables. Also, in Section \ref{sec:general_ssl}, we show that this framework can be employed to any additive State Space model, thereby we can use in a large class of engineering and statistical applications.

\subsection{Literature Review}

State space modeling for time series analysis represents a significant departure from traditional methods like Seasonal Autoregressive Integrated Moving Average (SARIMA). Various classical state space formulations, including the local level, linear trend, and basic structural model, have been extensively elucidated in sources such as \cite{Durbin_Koopman} and \cite{harvey_peters}. While many non-parametric methods, such as those depicted by \cite{boyd}, can extract unobserved components, the state space formulation provides a more comprehensive framework for time series analysis and forecasting. This approach, which relies on the Kalman filter for solving, offers robust solutions for modeling time series, including non-stationary ones. 

As discussed in \cite{Durbin_Koopman}, the estimation of the Kalman filter and smoother algorithm rely on nonlinear optimization whose performance (accuracy and solution time) is highly dependent on good initialization points. (as discussed in \cite{hiriart2001convex}).

Despite their efficacy, traditional state space frameworks face limitations, particularly in accommodating the demands of big data environments, as discussed in Section \ref{Introduction}. While applications of these models span diverse domains such as ecology \cite{ecological}, finance \cite{Mergner}, and engineering \cite{Sun}, methodological advancements have been sporadic, with few relevant exceptions concentrated on \cite{Koopman_Siem}, \cite{Mengheng}, \cite{Schiavoni_Koopman}, and \cite{Koopman_Lucas}.

Conversely, statistical learning methods have expanded significantly, with researchers pioneering innovative methodologies to bolster predictive accuracy and model interpretability. Widely embraced for their adeptness in handling big data, regularization methods like Lasso \cite{lasso} and Ridge Regression \cite{ridge} have emerged as stalwarts in combating overfitting issues inherent in linear regression models. The introduction of techniques such as the Elastic Net \cite{elasticnet} and Adaptive Lasso \cite{adalasso} further underscores the dynamic landscape of statistical learning. Recently, a variety of applications have utilized the statistical learning framework for tasks such as prediction, inference, and feature selection, as illustrated in \cite{Blanc,Cui}. For an larger collection of examples, see \cite{james2023statistical}.

Specifically in the time series domain, several statistical learning applications have been developed. For instance, \cite{Konzen} introduces the WLadaLASSO, a method employing weighted lag adaptive Lasso penalties in time series analysis. In \cite{Li_Chen}, the authors explore Lasso-based approaches for forecasting macroeconomic variables; they find that combining forecasts from Lasso-based methods with dynamic factor models further reduces forecast error, providing more interpretable and accurate predictions. Furthermore, in \cite{MEDEIROS2016255}, the authors thoroughly analyzed the asymptotic properties of the Adaptive Lasso within sparse, high-dimensional, linear time-series models. Notably, their results underscored the Adaptive Lasso's consistent selection of relevant variables as the volume of observations increased, indicative of model selection consistency.

We contend that a symbiotic fusion of statistical learning methodologies with the state-space framework holds immense potential. Such an interdisciplinary approach can invigorate and propel further development in the state-space literature on time series analysis, fostering a fertile ground for future research.

\pagebreak
\section{State-Space Learning Framework for Time Series Analysis}\label{meth}

In this section, we introduce and generalize a comprehensive framework that encompasses critical elements inherent to time series analysis, including stochastic trends, slopes, seasonality patterns, and disturbance
terms. Furthermore, we incorporate extensions for robust outlier detection, subset selection strategies, estimation procedures, and model selection criteria (all accomplished without necessitating a stationarity assumption), culminating in the formulation of an easily applicable extrapolation scheme. Additionally, our framework extends its utility beyond traditional time series data by providing a versatile model capable of accommodating diverse datasets, thereby enhancing its adaptability to various analytical contexts.

\subsection{Learning the Basic Structure of Time Series}\label{props}

A basic structural model can be formulated as

\begin{align}
    y_t &= \mu_t + \gamma_t + \varepsilon_t \label{eq1}\\
    \mu_{t+1} &= \mu_{t} + \nu_{t} + \xi_{t+1} \label{eq2}\\
    \nu_{t+1} &= \nu_{t} + \zeta_{t+1}\label{eq3}\\
    \gamma_{t+1} &= -\sum_{j=1}^{s-1} \gamma_{t+1 - j} + \omega_{t+1}\label{eq4},
\end{align} where $\mu_t$, $\nu_t$, $\gamma_t$ represents the stochastic trend, slope and seasonality respectively and $\mu_0$, $\nu_0$, $\gamma_1,\ldots,\gamma_s$ are the initial states.

In this Section, we will prove that this time series structure with level, trend, and seasonality can be written as a high-dimensional regression framework. This proof will be made separately for each component and then merged into one equation.

\begin{proposition}\label{proof_slope}
The slope component $\nu_{t+1} = \nu_{t} + \zeta_{t+1}$ can be formulated as $\nu_{t+1} = \nu_1 + \sum_{\tau=2}^{t+1}\zeta_{\tau}$.
 
\end{proposition}

\begin{proof}
We prove Proposition \ref{proof_slope} by induction. The statement $\nu_{t+1} = \nu_1 + \sum_{\tau=2}^{t+1}\zeta_{\tau}$ holds true for our base case $t=1$ since $\nu_2 = \nu_1 + \zeta_2$. Now, assuming the statement holds true for $t$, we show that it also holds true for $t+1$. For that,  we show that the induction hypothesis for $\nu_t $, i.e.,
\begin{equation}
\nu_t = \nu_1 + \sum_{\tau=2}^{t}\zeta_{\tau}, \label{eq:slope_hypothesis}
\end{equation}
can be confirmed for $t+1$ by using \eqref{eq:slope_hypothesis} in \eqref{eq3}, i.e., 
$$\nu_{t+1} = \nu_{t}+\zeta_{t+1} = \nu_1 + \sum_{\tau=2}^{t}\zeta_{\tau} + \zeta_{t+1} = \nu_1 + \sum_{\tau=2}^{t+1}\zeta_{\tau}.$$

\end{proof}

\begin{theorem}\label{proof_trend}
The trend component $\mu_{t+1} = \mu_{t} + \nu_{t} + \xi_{t+1}$ can be formulated as $$\mu_{t+1} = {\mu_1  + \sum_{\tau=2}^{t+1}\xi_\tau + t\,\nu_1 + \sum_{\tau=2}^{t}\bigl((t+1)-\tau\bigr)\zeta_\tau} $$.
 
\end{theorem}

\begin{proof}
We also prove it by induction. Our base case $\mu_{2} = \mu_{1} + \nu_{1} + \xi_2$ holds true for $t=2$ and we apply the induction hypothesis
$$\mu_t = \mu_1 + \sum_{\tau=2}^{t}\xi_\tau + (t-1)\nu_1 + \sum_{\tau=2}^{t-1}(t-\tau)\zeta_\tau$$
for $t\ge2$, in the recursion $\mu_{t+1} = {\mu_{t}} + \nu_{t} + \xi_{t+1}$ and obtain
$$\mu_{t+1} = {\mu_1 + \sum_{\tau=2}^{t}\xi_\tau + (t-1)\nu_1 + \sum_{\tau=2}^{t-1}(t-\tau)\zeta_\tau} + \nu_{t} + \xi_{t+1}$$ Now, using Proposition \eqref{proof_slope}, we replace $\nu_{t}$ obtain
$$\mu_{t+1} = {\mu_1 + \sum_{\tau=2}^{t}\xi_\tau + (t-1)\nu_1 + \sum_{\tau=2}^{t-1}(t-\tau)\zeta_\tau}+ \nu_1 + \sum_{\tau=2}^{t}\zeta_{\tau} + \xi_{t+1}
$$
and, consequentially,
$$\mu_{t+1} = {\mu_1 + \sum_{\tau=2}^{t+1}\xi_\tau + ((t+1)-1)\nu_1 + \sum_{\tau=2}^{(t+1)-1}((t+1)-\tau)\zeta_\tau} 
$$

\end{proof}

\begin{theorem}\label{proof_seas}
The seasonal component $\gamma_{t+1} = -\sum_{j=1}^{s-1} \gamma_{t+1 - j} + \omega_{t+1}$ can be formulated as

$$\gamma_{t+1} = \gamma_{{t+1}-ks}+\sum_{j=0}^{k-1}(\omega_{{t+1}-js} -  \omega_{t-js}), \quad \forall t \in \{1,\cdots,T-1\}, k \in \left\{1,\ldots,\floor*{\frac{t}{s}}\right\},$$ given that $\floor*{\cdot}$ is the floor operator.
 
\end{theorem}

\begin{proof}

We first define our induction hypothesis as

$$\gamma_{t+1} = \gamma_{{t+1}-ks}+\sum_{j=0}^{k-1}(\omega_{{t+1}-js} -  \omega_{t-js}), \quad \forall t \in \{1,\cdots,T-1\}, k \in \left\{1,\ldots,\floor*{\frac{t}{s}}\right\}.$$

To prove that our base case ($k=1$) holds,  we rewrite Equation \ref{eq4} and apply it for ${t+1}$ and $t$. 
\begin{align}
    \sum_{j=1}^{s}\gamma_{t+2-j} = \omega_{t+1} \label{eq5}\\
    \sum_{j=1}^{s}\gamma_{t+1-j}=\omega_{t} \label{eq6}
\end{align}

Subtracting \ref{eq6} from \ref{eq5}, we obtain the seasonal recursion
\begin{align}
    \gamma_{{t+1}} = \gamma_{{t+1}-s} + \omega_{t+1}-\omega_{t}, \quad\forall t\in\{s,\ldots,T-1\} \label{mod_seas}.
\end{align}

Now, we prove that if the hypothesis holds for $k=1$,
it also holds for any $ k \in \left\{1,\ldots,\floor*{\frac{t}{s}}\right\}$. For that, we replace $\gamma_{{t+1}-ks}$ in
$$\gamma_{t+1} = \gamma_{{t+1}-ks}+\sum_{j=0}^{k-1}(\omega_{{t+1}-js} -  \omega_{t-js})$$
by its seasonal recursion $\gamma_{{t+1}-ks} = \gamma_{{t+1}-(k+1)s} + \omega_{{t+1}-ks} - \omega_{{t}-ks}$ and obtain

$$\gamma_{t+1} = \gamma_{{t+1}-(k+1)s} + \omega_{{t+1}-ks} - \omega_{{t}-ks}+\sum_{j=0}^{k-1}(\omega_{{t+1}-js} -  \omega_{t-js}).$$

Therefore, 
$$\gamma_{t+1} = \gamma_{{t+1}-(k+1)s}+\sum_{j=0}^{k}(\omega_{{t+1}-js} -  \omega_{t-js}).$$

Now, assuming $k=\floor*{\frac{t}{s}}$, i.e., its largest possible $k$ value for each time $t+1$, we define $m_{t+1} = {t+1} - s\floor*{\frac{t}{s}}$. For instance, if we have a monthly frequency and $s=12$, then $m_{t+1}$ can be interpreted as the first occurrence of the month associated to time ${t+1}$. Hence, we define our seasonal component as
$$\gamma_{t+1} = \gamma_{m_{t+1}} + \sum_{j=0}^{\floor*{{t}/{s}}-1}(\omega_{{t+1}-js} -  \omega_{t-js}).$$

For didactic purposes, we define $\tau = t+1-js$ reformulate it as
$$\gamma_{t+1} = \gamma_{m_{t+1}} + \sum_{\tau \in M_{t+1}}(\omega_{{\tau}} -  \omega_{\tau-1}),$$
where $M_{t+1} = \{{t+1}-js\}_{j=0}^{\floor*{{t}/{s}}-1} = \{m_{t+1}+s,m_{t+1}+2s,\ldots, t+1-2s,t+1-s,t+1\}$. Formally speaking, $M_{t} = \{\tau \in \{m_{t}+s,\ldots,t\}\,|\, m_{t}=m_\tau\}.$

\end{proof}

Finally, we obtain our final formulation, where the basic structural model is fully represented by a high-dimensional regression.

\begin{corollary}\label{coro}
The basic structural model (\ref{eq1}-\ref{eq4}) is equivalent to the high dimensional regression 
\begin{align}
    y_1 &= \mu_1 + \gamma_1 + \varepsilon_1 \label{cor:t1}\\
    y_2 &= \mu_1 + \xi_2 + \nu_1 + \gamma_2 + \varepsilon_2 \\
    y_t &= \mu_1 + \sum_{\tau=2}^{t}\xi_\tau  + (t-1)\nu_1 + \sum_{\tau=2}^{t-1}(t-\tau)\zeta_\tau + \gamma_{m_t} + \varepsilon_t, \quad \forall t = \{3, \ldots, s\} \label{cor_t2}\\
    y_t &= \mu_1  + \sum_{\tau=2}^{t}\xi_\tau + (t-1)\nu_1 + \sum_{\tau=2}^{t-1}(t-\tau)\zeta_\tau + \gamma_{m_{t}} + \sum_{\tau \in M_{t}}(\omega_{{\tau}} -  \omega_{\tau-1})+  \varepsilon_t, \quad \forall t=s+1,\ldots,T \label{proposition}
\end{align}
\end{corollary}

\begin{proof}
    First, \eqref{cor:t1} is the same as \eqref{eq1} for $t=1$. Second, we prove \eqref{cor_t2} by replacing in \eqref{eq1}  the definition of $\mu_t$ in \eqref{eq2}, for $t=2$. Then, we have $y_2 = \mu_2 + \gamma_2 + \varepsilon_2 = \mu_1 + \nu_1 + \xi_2 + \gamma_2 + \varepsilon_2.$
    Finally, we prove the corollary for $t \ge 2$ by replacing in the equation $y_t = \mu_t + \gamma_t + \varepsilon_t$ the representations of the trend $\mu_t$ and seasonality $\gamma_t$ obtained in Theorem \ref{proof_trend} and in Theorem \ref{proof_seas}, respectively. 
\end{proof}

\subsection{Interpreting the underlying regressors}

For an easier understanding of the underlying regressors, we rewrite Corollary \ref{coro} as
\begin{align}
y_t &= \mu_1 + (t-1)\nu_1 + \sum_{k=1}^s \mathbb{I}_{\{m_t\}}(k)\,\gamma_k \nonumber\\ &+ \sum_{\tau=2}^{T}\left[\mathbb{I}_{[1,t]}(\tau) \, \xi_\tau  + (t-\tau)^+ \,\zeta_\tau +  \left(\mathbb{I}_{M_t}(\tau) - \mathbb{I}_{M_{t+1}}(\tau)\right)\,\omega_{{\tau}}\right]+  \varepsilon_t,\label{eq:ident}
\end{align} where 
$x^+=\max(0,x)$ and
$\mathbb{I}_A(x)$ is an indicator function that values 1 when $x \in A$ and 0 otherwise. 
For a better visualization of the model as a regression, let $Y = (y_1,\ldots,y_T)^\top$, $\varepsilon = (\varepsilon_1,\ldots,\varepsilon_T)^\top$  and $\Theta=(\mu_1,\xi,\nu_1,\zeta, \gamma_1, \ldots, \gamma_s, \omega)^\top$, where $\xi = \xi_2,\ldots,\xi_T$, $\zeta = \zeta_2,\ldots,\zeta_T$ and $\omega = \omega_s,\ldots,\omega_T$, then we can represent it as $Y = X\,\Theta + \varepsilon$, where $X$ is the regressor matrix. For instance, for $T=5$ and $s=2$, the regressor matrix $X$ is defined as 
\begin{equation}\label{Eq:matrix1}
    \begin{blockarray}{cccccccccccccccc}
    \mu_1  & \xi_2 & \xi_3 & \xi_4 & \xi_5 & \nu_1 & \zeta_2  & \zeta_3 & \zeta_4 & \gamma_1 & \gamma_2  & \omega_2 & \omega_3 & \omega_4 & \omega_5\\
    \begin{block}{(cccccccccccccccc)}
    1 & 0 & 0 & 0 & 0  & 0 & 0 & 0 & 0 & 1 & 0 & 0 & 0 & 0 & 0 \\
    1 & 1 & 0 & 0 & 0  & 1 & 0 & 0 & 0 & 0 & 1 & 0 & 0 & 0 & 0  \\
    1 & 1 & 1 & 0 & 0  & 2 & 1 & 0 & 0 & 1 & 0 & -1 & 1 & 0 & 0 \\
    1 & 1 & 1 & 1 & 0  & 3 & 2 & 1 & 0 & 0 & 1 & 0 & -1 & 1 & 0 \\
    1 & 1 & 1 & 1 & 1  & 4 & 3 & 2 & 1 & 1 & 0 & -1 & 1 & -1 & 1 \\
    \end{block}
    \end{blockarray}.
    \end{equation}

We now provide a quick interpretation of the components of the proposed high-dimensional regression framework (level, slope, and seasonality). The initial state of the level component, represented by $\mu_1$, corresponds to the model's intercept. The regressors associated with the innovations $\xi_t$ will assume value 1 when $t \ge \tau$ and 0 otherwise. These regressors can be conceptualized as a series of step functions, each starting at distinct times, aiming to capture level changes within the time series. It's important to highlight that a regularized model aims to parsimoniously select the pertinent step functions to elucidate the underlying dynamics of the time series. 

For the slope component, it's important to recognize that the initial state $\nu_1$ is the coefficient of a conventional linear trend regressor with values spanning from 0 to $T-1$. Complementing this, the coefficients $\zeta_t$ refer to regressors resembling rectified linear unit (ReLU), initiating at each timestamp. Together, they form a piecewise linear function adept at capturing changes in trends across time.

\begin{figure}[H]
    \centering
    \begin{subfigure}[b]{0.4\textwidth}
    \includegraphics[width=\textwidth]{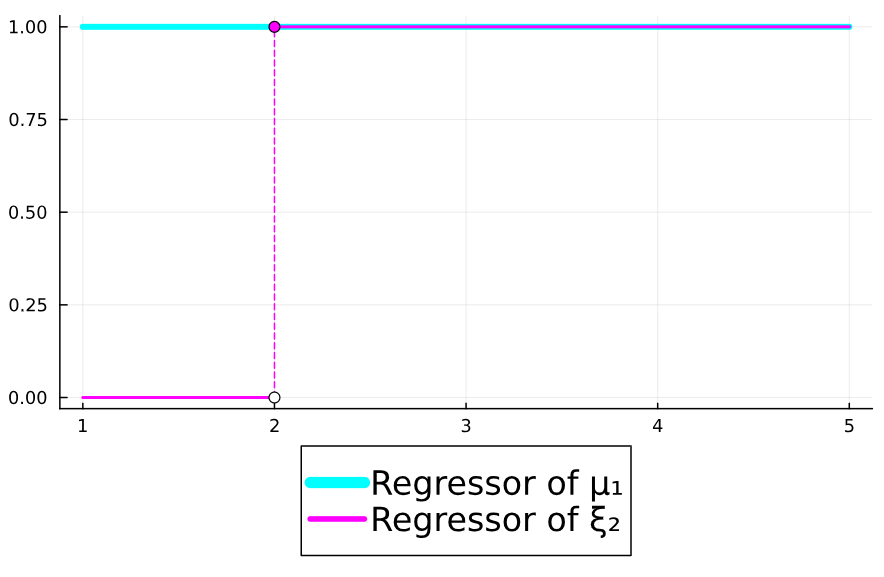}
    \caption{}
    \label{fig:sub1}
    \end{subfigure}
    \hfill
    \begin{subfigure}[b]{0.4\textwidth}
    \includegraphics[width=\textwidth]{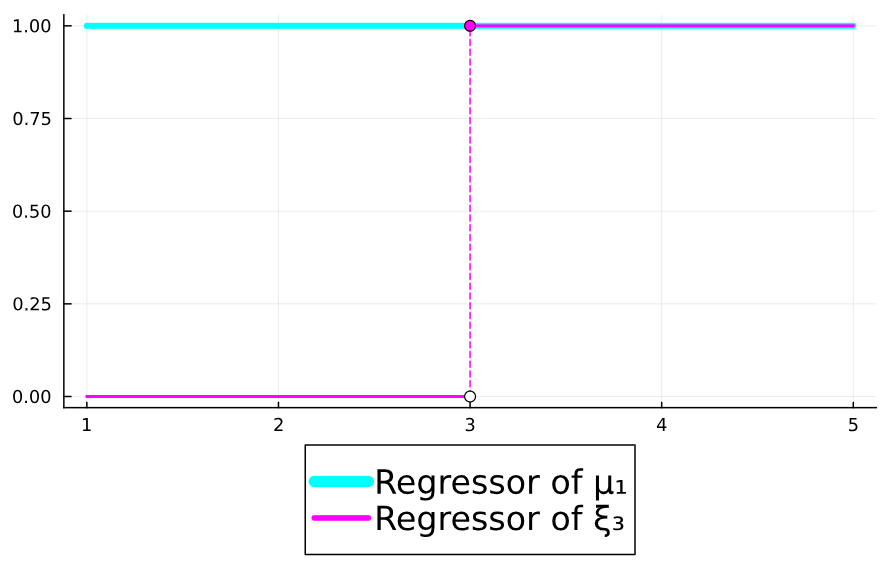}
    \caption{}
    \label{fig:sub2}
    \end{subfigure}
    
    \vspace{0.5cm}
    
    \begin{subfigure}[b]{0.4\textwidth}
    \includegraphics[width=\textwidth]{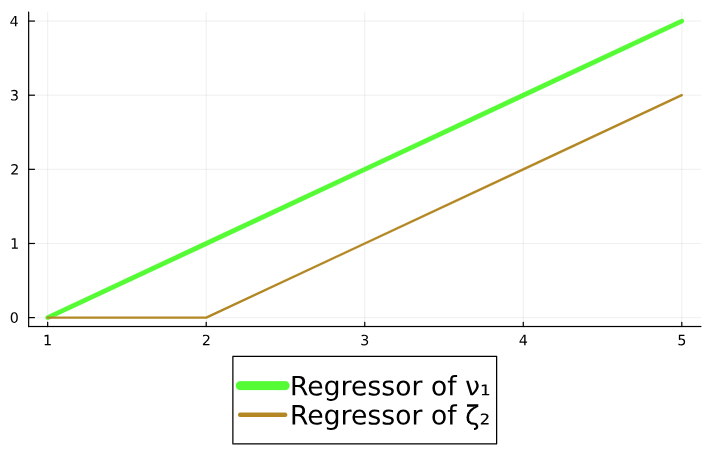}
    \caption{}
    \label{fig:sub4}
    \end{subfigure}
    \hfill
    \begin{subfigure}[b]{0.4\textwidth}
    \includegraphics[width=\textwidth]{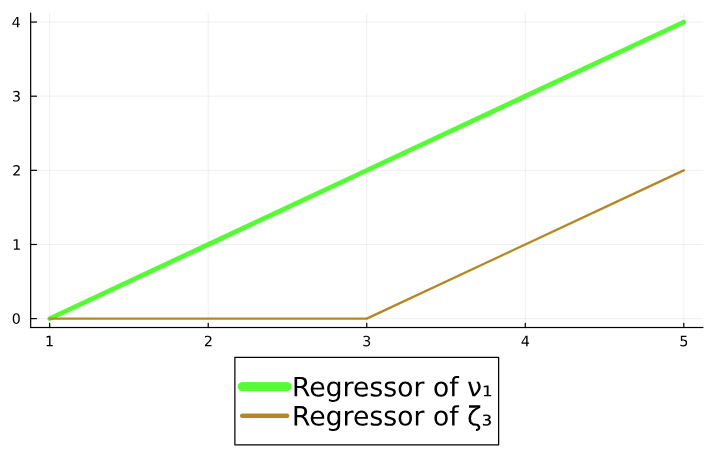}
    \caption{}
    \label{fig:sub5}
    \end{subfigure}
    \caption{In this example we see in Figures \ref{fig:sub1}, \ref{fig:sub2} the components $\mu_1 + \sum_{\tau=2}^{t}\xi_\tau$ and in Figures \ref{fig:sub4}, \ref{fig:sub5} the components $(t-1)\nu_1 + \sum_{\tau=2}^{t-1}(t-\tau)\zeta_\tau$.}
    \label{fig:figure_comp}
    \end{figure}

In Figure \ref{fig:figure_comp}, we present an illustrative example of these components for a time series with 5 observations. The initial states $\gamma_1,\ldots,\gamma_s$ of the seasonal component can be likened to seasonal dummy regressors, representing the deterministic part of the seasonality. Meanwhile, the innovations $\omega_t=\sum_{j=1}^s \gamma_{t+1-j}$ can be conceptualized as moving sums of the stochastic seasonal components spanning an entire season. It's worth noting that in a regularized regression model, there's a tendency for $\omega_t$ to shrink towards zero. This occurs because seasonal variations tend to cancel out when summed over a whole season, such as twelve months, leading to a diminished impact on the overall model.

\subsection{Estimation Procedure, Model Selection and Component Extraction}\label{estimation_p}

We introduce an estimation procedure featuring adaptive regularization, akin to the Adaptive Lasso method \citep{adalasso}. However, our approach incorporates a distinctive adaptation specifically designed to address the nuances of the present problem. To obtain estimates for $\Theta$, we first solve
\begin{align}
    \underset{\theta_0,\theta}{\text{minimize }} & \frac{1}{|\mathcal{T}|}\sum_{t \in \mathcal{T}} \left[y_t - \hat{y}_t(\theta_0,\theta)\right]^2 + \lambda\left[(1-\alpha)\frac{1}{2}  \|\theta\|_2^2+\alpha \|\theta\|_1\right] \label{adalasso_equation1},
\end{align}
where $\mathcal{T}=\{1,\ldots,T\}$, $\theta_0 = (\mu_1, \nu_1, \gamma_1,\ldots,\gamma_s)$ are the initial states  and $\theta = (\xi_2,\ldots,\xi_T, \zeta_2,\ldots,\zeta_T,\\\omega_s, \ldots,\omega_T)$. 

Except for the initial states, note that the regularization manifests as a penalty $\lambda$ applied to the convex combination of the L1 and L2 norms of $\theta$ with weights $\alpha$ and $(1-\alpha)$, respectively. It is crucial to underscore the deliberate exclusion of regularization on the initial states during this phase. This intentional omission allows us to effectively identify and obtain a parsimonious impact of the innovations while preserving the key deterministic components of the model: intercept, a linear deterministic trend, and the seasonal dummies. 
For a fixed $\alpha$, we solve \eqref{adalasso_equation1} for several values of $\lambda$ and chooses the one the minimizes an information criteria, such as AIC or BIC.

In the second step, we develop adaptive estimation procedure 
\begin{align}
 \begin{split}
\underset{\theta_0,\theta}{\text{minimize }}  \frac{1}{|\mathcal{T}|}\sum_{t\in\mathcal{T}} \left[y_t - \hat{y}_t(\theta_0,\theta)\right]^2+ 
    \lambda \Biggl[ & w_{\xi}\left( (1-\alpha)\frac{1}{2}    \|\xi\|_2^2 + \alpha    \|\xi\|_1 \right) +\\ 
    & w_{\zeta}\left((1-\alpha)\frac{1}{2}    \|\zeta\|_2^2 + \alpha    \|\zeta\|_1 \right) +\\
    & w_{\omega}\left((1-\alpha)\frac{1}{2}    \|\omega\|_2^2 + \alpha    \|\omega\|_1 \right) + \\ 
     & w_{\nu_1} \left((1-\alpha)\frac{1}{2}    \|\nu_1\|_2^2 + \alpha    \|\nu_1\|_1 \right) + \\
     & \sum_{i=1}^s w_{\gamma_i} \left((1-\alpha)\frac{1}{2}    \|\gamma_i\|_2^2 + \alpha    \|\gamma_i\|_1 \right) \Biggr],  \\
 \end{split}
\label{adalasso_equation2}
\end{align}
where we introduce different penalty weights $w_{\xi},w_{\zeta},w_{\omega},w_{\nu_1},w_{\gamma_1},\ldots,w_{\gamma_s} $ for each regression coefficient.
To define these weights, let $\hat\theta_0 = (\hat\mu_1, \hat\nu_1, \hat\gamma_1,\ldots,\hat\gamma_s)$ and $\hat\theta = (\hat\xi_2,\ldots,\hat\xi_T, \hat\zeta_2,\ldots,\hat\zeta_T, \hat\omega_s, \ldots,\hat\omega_T)$ denote the solutions of \eqref{adalasso_equation1}. Next, we calculate estimates for the standard deviation of each innovation term $\hat\sigma_\kappa=\|\kappa\|_2, \forall \kappa \in \{\xi,\zeta,\omega\}$ and derive the associated penalty weights 
$w_{\kappa} = \frac{1}{\epsilon +\hat{\sigma}_{\kappa}}, \forall \kappa \in \{\xi,\zeta,\omega\}$ for the second step of our adaptive procedure. Likewise, we introduce penalty weights for $w_{\nu_1} = \frac{1}{\epsilon + |\hat\nu_1|}$ and $w_{\gamma_i} = \frac{1}{\epsilon + |\hat\gamma_i|}, \forall i=1,\ldots,s$. The term $\epsilon$ represents a non-negative parameter that handles the case where the estimated coefficients of the first step are 0 \cite{Ballout}. In our numerical experiments, we set $\epsilon=0.05$. We contend that incorporating a penalty in the second step for the linear trend and seasonal dummies facilitates robust estimation and automated selection of these structures.

For the same $\alpha$ fixed in the first step, we choose the regularization penalty $\lambda$ such as to minimize the information criteria of choice (e.g., AIC or BIC). Now, let $\theta_0 = (\mu_1, \nu_1, \gamma_1,\ldots,\gamma_s)$  and $\theta = (\xi_2,\ldots,\xi_T, \zeta_2,\ldots,\zeta_T, \omega_s, \ldots,\omega_T)$
denote the solution of \eqref{adalasso_equation2} for the selected $\alpha$ and $\lambda$. Then, we can reconstruct the components using the Theorems in Section \ref{props}. Formally speaking:
\begin{align}
    \nu_t &= \nu_1 + \sum_{\tau=2}^t\zeta_\tau \label{eq:comp1}\\
    \mu_t &= \mu_1 + \sum_{\tau=2}^{t}\xi_\tau + (t-1)\nu_1 + \sum_{\tau=2}^{t-1}(t-\tau)\zeta_\tau\label{eq:comp2}\\ 
    \gamma_{t} &= \gamma_{m_{t}} + \sum_{\tau \in M_{t}}(\omega_{{\tau}} -  \omega_{\tau-1}).\label{eq:comp3}
\end{align}

The extraction of model components constitutes a pivotal phase in structural time series analysis, facilitating the discernment and exploration of trends and patterns that may elude direct observation in raw data. We illustrate component extraction capability of the model and compare it with a traditional benchmark. Specifically, we compare the smoothed components of a Gaussian benchmark estimated via Kalman filter against the SSL ($AIC, \alpha=0.1$) model. This comparative analysis is performed on three diverse time series drawn from the M4 dataset, see \cite{m4competitorsguide}. 

\begin{figure}[H]
    \centering
    \begin{subfigure}[b]{0.32\textwidth}
        \includegraphics[width=\textwidth]{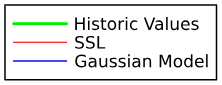}
    \end{subfigure}
    \hfill
    \begin{subfigure}[b]{0.32\textwidth}
        \includegraphics[width=\textwidth]{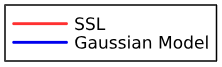}
    \end{subfigure}
    \hfill
    \begin{subfigure}[b]{0.32\textwidth}
        \includegraphics[width=\textwidth]{img/lab2.png}
    \end{subfigure}
    \vskip\baselineskip
    \begin{subfigure}[b]{0.32\textwidth}
        \includegraphics[width=\textwidth]{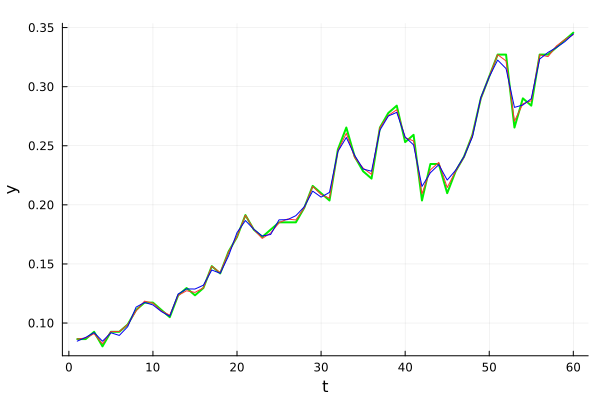}
    \end{subfigure}
    \hfill
    \begin{subfigure}[b]{0.32\textwidth}
        \includegraphics[width=\textwidth]{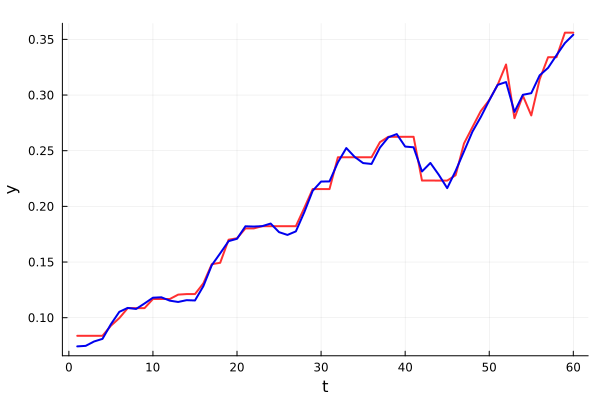}
    \end{subfigure}
    \hfill
    \begin{subfigure}[b]{0.32\textwidth}
        \includegraphics[width=\textwidth]{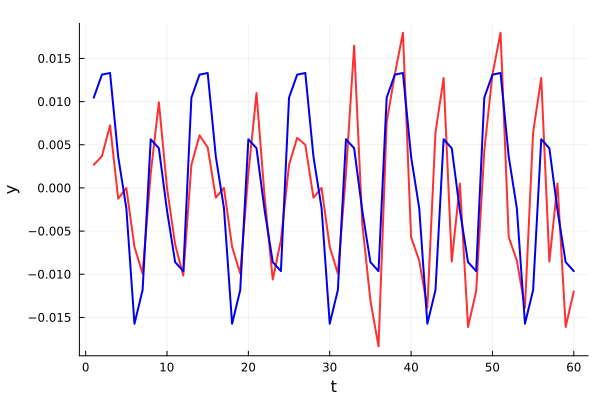}
    \end{subfigure}
    \vskip\baselineskip
    \begin{subfigure}[b]{0.32\textwidth}
        \includegraphics[width=\textwidth]{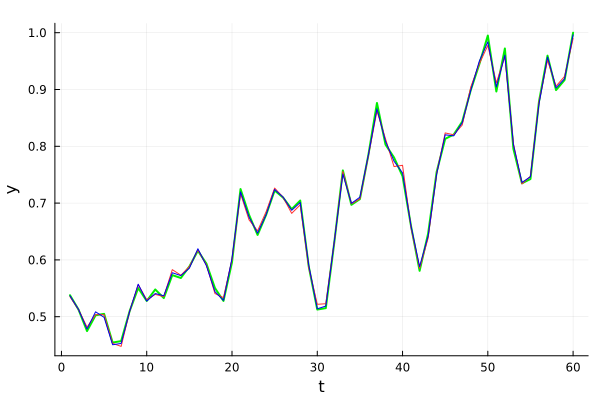}
    \end{subfigure}
    \hfill
    \begin{subfigure}[b]{0.32\textwidth}
        \includegraphics[width=\textwidth]{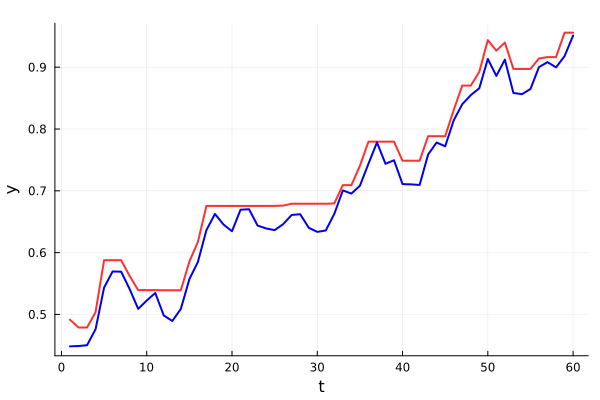}
    \end{subfigure}
    \hfill
    \begin{subfigure}[b]{0.32\textwidth}
        \includegraphics[width=\textwidth]{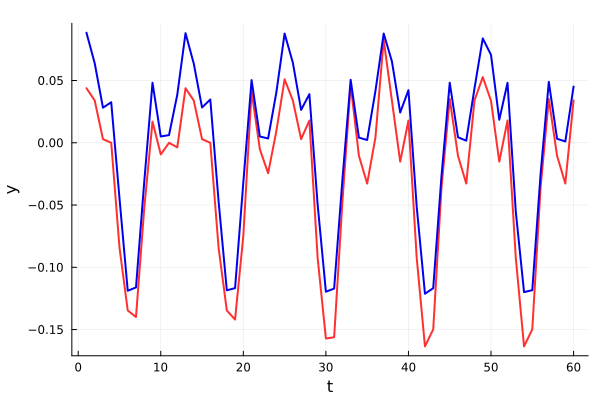}
    \end{subfigure}
    \vskip\baselineskip
    \begin{subfigure}[b]{0.32\textwidth}
        \includegraphics[width=\textwidth]{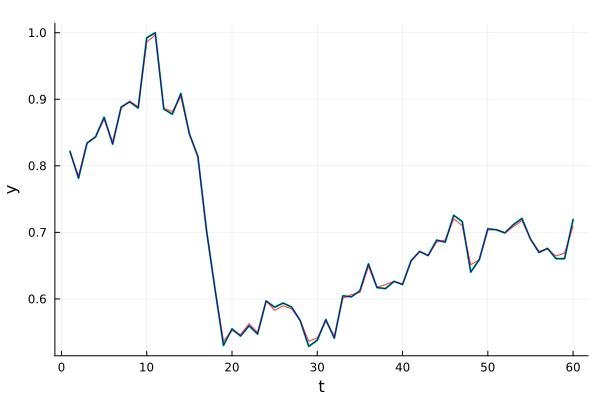}
        \caption{Fitted Models}
    \end{subfigure}
    \hfill
    \begin{subfigure}[b]{0.32\textwidth}
        \includegraphics[width=\textwidth]{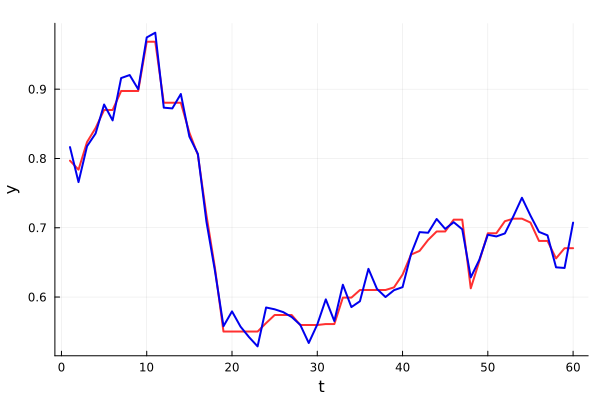}
        \caption{Trend Components}
    \end{subfigure}
    \hfill
    \begin{subfigure}[b]{0.32\textwidth}
        \includegraphics[width=\textwidth]{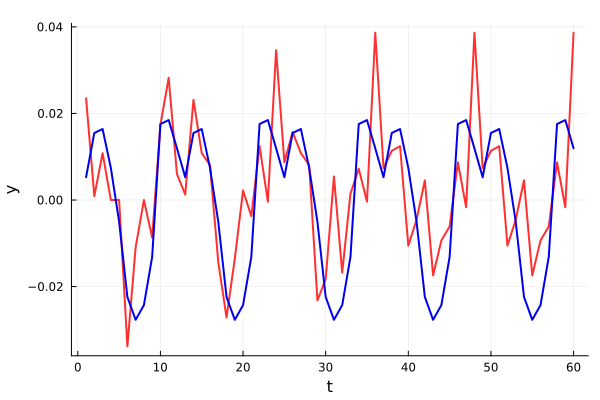}
        \caption{Seasonality Components}
    \end{subfigure}
    \hfill
    \caption{Fitted time series and separated components (each row represents one time series) for both the gaussian and the SSL $(\alpha=0.1-AIC)$ models.}
    \label{comps_}
\end{figure}

Our findings, graphically depicted in Figure \ref{comps_}, reveal notable similarities in the fits produced by both models. Note that the results in for the SSL model presents a less noisy pattern (mainly in the trend component), which indicates that our proposed methodology presents more parsimonious results.

\subsection{Handling Missing Values via Interpolation}

The proposed SSL framework has the capability to handle missing values within time-series data through a slight adjustment in the estimation procedure. Specifically, in both Equations \eqref{adalasso_equation1} and \eqref{adalasso_equation2}, we define the $\mathcal{T} \in \{1,\ldots,T\}$ to be the set of timestamps with valid observations. It is noteworthy that the fitted values of the estimated time series can be employed to populate missing values at those time stamps where observations are absent, as these fitted values serve as interpolations of the underlying time series dynamics.
Formally speaking, the interpolated values for $t\notin \mathcal{T}$ is given by $\hat y_t = \mu_t + \gamma_t$, where $\mu_t$ and $\gamma_t$ are computed as in (\ref{eq:comp1}-\ref{eq:comp3}), using the estimate $\Theta$ obtained as a solution of \eqref{adalasso_equation2}.

Allow us to elucidate instances of time series exhibiting lacunae in data, alongside the interpolated outcomes for these instances, as facilitated by both the proposed State Space Learning framework and its Gaussian analogue. In this examination, three authentic time series were subjected to the removal of designated observations, thereby categorizing them as absent values, to evaluate the efficacy of our proposed framework. Evidently, our analysis demonstrates that both models yield robust outcomes for the missing observations, as evidenced by comparison with the original values, which were not considered throughout the estimation procedure.

\begin{figure}[H]
    \centering
    \begin{subfigure}[b]{0.32\textwidth}
        \centering
        \includegraphics[width=\textwidth]{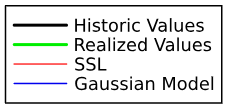}
    \end{subfigure}
    \hfill
    \begin{subfigure}[b]{0.32\textwidth}
        \centering
        \includegraphics[width=\textwidth]{img/lab_pred.png}
    \end{subfigure}
    \hfill
    \vskip\baselineskip
    \begin{subfigure}[b]{0.44\textwidth}
        \centering
        \includegraphics[width=\textwidth]{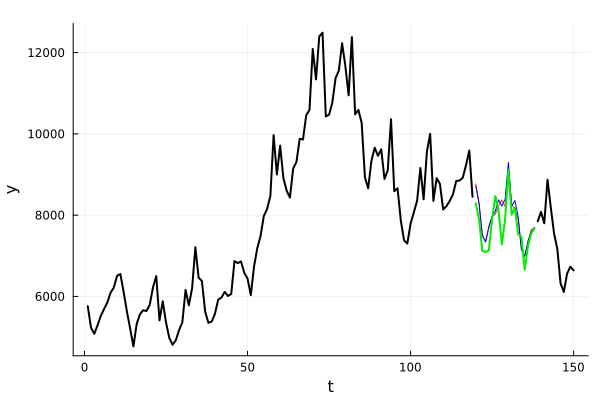}
    \end{subfigure}
    \hfill
    \begin{subfigure}[b]{0.44\textwidth}
        \centering
        \includegraphics[width=\textwidth]{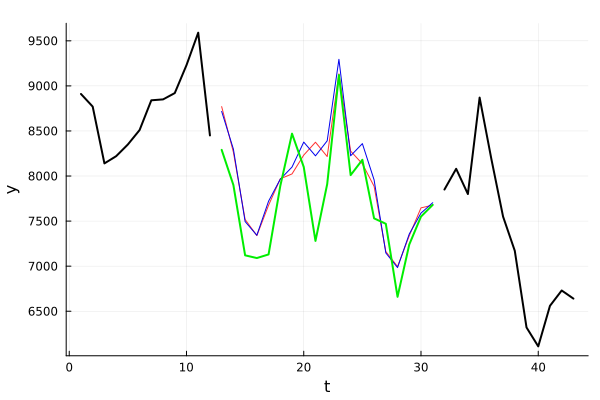}
    \end{subfigure}
    \hfill
    \vskip\baselineskip
    \begin{subfigure}[b]{0.42\textwidth}
        \includegraphics[width=\textwidth]{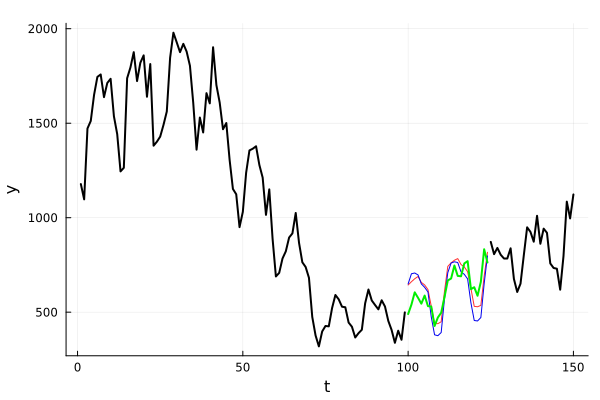}
    \end{subfigure}
    \hfill
    \begin{subfigure}[b]{0.42\textwidth}
        \includegraphics[width=\textwidth]{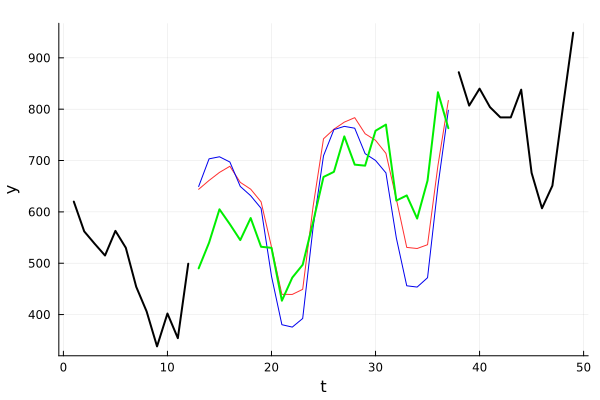}
    \end{subfigure}
    \hfill
    \vskip\baselineskip
    \begin{subfigure}[b]{0.42\textwidth}
        \includegraphics[width=\textwidth]{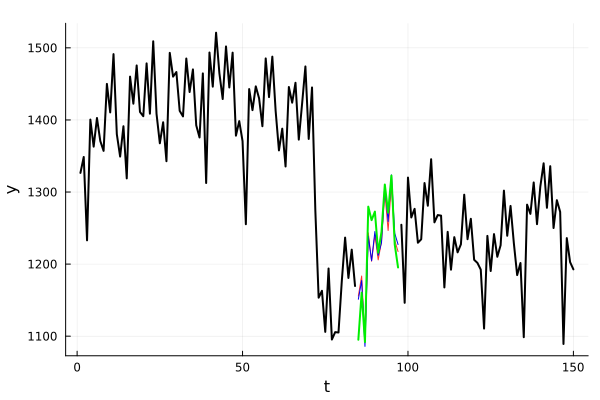}
        \caption{Filled missing values of both the gaussian and the SSL $(\alpha=0.1-AIC)$ models in three different time series.}
    \end{subfigure}
    \hfill
    \begin{subfigure}[b]{0.42\textwidth}
        \includegraphics[width=\textwidth]{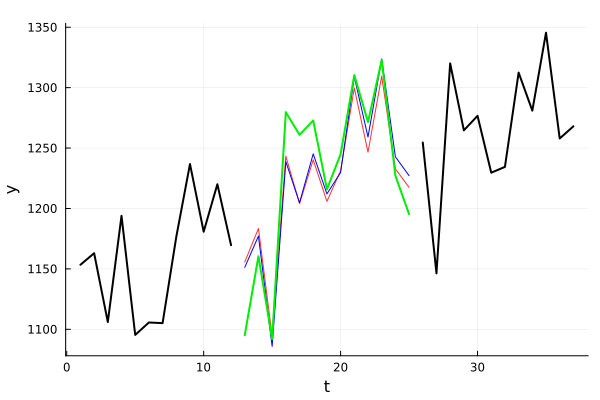}
        \caption{Zoom of the filled missing values of both the gaussian and the SSL $(\alpha=0.1-AIC)$ models in three different time series from the M4 dataset.}
    \end{subfigure}
    \hfill
    \caption{Model Forecasts}
    \label{missing_values}
\end{figure}

\subsection{Forecasting via Extrapolation}

A robust forecasting scheme within a time series analysis framework is paramount for informed decision-making across various domains, including finance, economics, and climate science. Accurate forecasts enable stakeholders to anticipate future trends, allocate resources efficiently, and mitigate potential risks. Moreover, reliable forecasts facilitate proactive planning, aiding businesses in optimizing their decisions. Therefore, the development and implementation of effective forecasting methodologies are essential for fostering resilience and sustainability in dynamic environments. We propose an intuitive forecasting scheme for the SSL framework, relying on the extrapolation of each component of the high-dimensional regression matrix and the utilization of estimated coefficients. This approach not only simplifies the forecasting process but also ensures accuracy and adaptability, making it a valuable tool for addressing the forecasting needs of complex systems.
\begin{proposition}\label{for_prop}
Assuming that the values of $\Theta=(\mu_1,\xi_2,\ldots,\xi_T,\nu_1,\zeta_2,\ldots,\zeta_5, \gamma_1, \ldots, \gamma_s, \omega_s,\ldots,\omega_T)^\top$ are given and $E[\varepsilon_t|\Theta]=0$,$E[\xi_t|\Theta]=0$, $E[\zeta_t|\Theta]=0$, and $E[\omega_t|\Theta]=0$ for every $t >T$, then a forecast for $t > T$ is defined as 

\begin{align}
\hat{y}_{t} = \mu_1 + (t-1)\nu_1 + & \sum_{\tau=2}^{T-1}(t-\tau)\zeta_\tau + \sum_{\tau=2}^{T}\xi_\tau + \gamma_{m_t}  +\sum_{\tau \in M_{T}}(\omega_{{\tau}} -  \omega_{\tau-1}), \quad \forall t > T
\end{align}\label{for_eq}
\end{proposition}

\begin{proof}

Given the Equation \ref{eq:ident} and that $\Theta$ is a given set of coefficients $(\mu_1,\xi_2,\ldots,\xi_T,\nu_1,\zeta_2,\ldots,\zeta_5, \\ \gamma_1, \ldots, \gamma_s, \omega_s,\ldots,\omega_T)^\top$, the value of $y_t$ for $t > T$ is defined as:
\begin{align}
 y_t = \mu_1 + (t-1)\nu_1 + \sum_{\tau=2}^{t-1}(t-\tau)\zeta_\tau + \sum_{\tau=2}^{t}\xi_\tau + \gamma_{m_t}+\sum_{\tau \in M_{t}}(\omega_{{\tau}} -  \omega_{\tau-1}) +  \varepsilon_t.
\end{align}Therefore, the forecast will be calculated as:
   \begin{align}
   \begin{split}
 \hat{y}_{t} = E[y_t|\Theta] = 
 \mu_1 + (t-1)\nu_1 + & \sum_{\tau=2}^{T}(t-\tau)E[\zeta_\tau|\Theta]+ \sum_{\tau=T+1}^{t-1}(t-\tau)E[\zeta_\tau|\Theta] \\
 &+ \sum_{\tau=2}^{T}E[\xi_\tau|\Theta] + \sum_{\tau=T+1}^{t}E[\xi_\tau|\Theta] + \gamma_{m_t} \\
 &+\sum_{\tau \in M_T}(E[\omega_{\tau}|\Theta] -  E[\omega_{\tau-1}|\Theta]) \\
 &+\sum_{\tau \in M_t \setminus M_T}(E[\omega_{\tau}|\Theta] -  E[\omega_{\tau-1}|\Theta]) +  E[\varepsilon_t],
 \end{split}
\end{align}
where the term $M_t \setminus M_T$ specifies the set of elements that belongs to $M_t$ and do not belongs to $M_T$.

Now, for each innovation $\kappa \in \{\xi,\zeta,\omega\}$, we have $E[\kappa_{\tau}|\Theta] = \kappa_\tau, \forall \tau \le T$ and $E[\kappa_{\tau}|\Theta] = 0, \forall \tau > T$, thereby
$$\hat{y}_t = \mu_1 + (t-1)\nu_1 + \sum_{\tau=2}^{T-1}(t-\tau)\zeta_\tau + \sum_{\tau=2}^{T}\xi_\tau + \gamma_{m_t}+\sum_{\tau \in M_{t}}(\omega_{{\tau}} -  \omega_{\tau-1}).$$

\end{proof}

For a better visualization, we illustrate the forecasting process by showing the process of fit and forecast (for the case where T = 5 and s = 2) and a forecasting process of two steps ahead.
The fit process will estimate the $\theta_0 = \mu_1$ and $\theta$ = [$\nu_1$, $\zeta_1,\zeta_2,\zeta_3,\zeta_4$, $\xi_1, \xi_2, \xi_3,\xi_4$, $\gamma_1, \gamma_2$, $\omega_2,\omega_3,\omega_4$] parameters.

\begin{equation}\label{Eq:matrix0}
\begin{blockarray}{c}
    y \\
    \begin{block}{(c)}
    y_1\\
    y_2\\
    y_3\\
    y_4\\
    y_5\\
    \end{block}
    \end{blockarray} = 
    \begin{blockarray}{ccccccccccccccc}
    \mu_1 & \xi_2 & \xi_3 & \xi_4 & \xi_5 & \nu_1 & \zeta_2 & \zeta_3 & \zeta_4 & \gamma_1 & \gamma_2 & \omega_2 & \omega_3 & \omega_4 & \omega_5\\
    \begin{block}{(ccccccccccccccc)}
    1 & 0 & 0 & 0 & 0  & 0 & 0 & 0 & 0  & 1 & 0 & 0 & 0 & 0 & 0  \\
    1 & 1 & 0 & 0 & 0 & 1 & 0 & 0 & 0  & 0 & 1 & 0 & 0 & 0 & 0   \\
    1 & 1 & 1 & 0 & 0 & 2 & 1 & 0  & 0 & 1 & 0 & -1 & 1 & 0 & 0  \\
    1 & 1 & 1 & 1 & 0 & 3 & 2 & 1 & 0 & 0 & 1 & 0 & -1 & 1 & 0 \\
    1 & 1 & 1 & 1 & 1 & 4 & 3 & 2  & 1 & 1 & 0 & -1 & 1 & -1 & 1 \\
    \end{block}
    \end{blockarray}
    \cdot
    \begin{blockarray}{c}
    \begin{block}{(c)}
    \theta_0\\
    \theta \\
    \end{block}
    \end{blockarray}
    \end{equation}

Finally, the forecast will be obtained by employing the estimated $\hat{\theta}_0$ and $\hat{\theta}$ in the following equation:

\begin{equation}\label{Eq:matrix2}
\begin{blockarray}{c}
    y \\
    \begin{block}{(c)}
    \hat{y}_6\\
    \hat{y}_7\\
    \end{block}
    \end{blockarray} = 
    \begin{blockarray}{cccccccccccccccc}
    \mu_1 & \xi_2 & \xi_3 & \xi_4 & \xi_5 & \nu_1 & \zeta_2 & \zeta_3  & \zeta_4 & \gamma_1 & \gamma_2 & \omega_2 & \omega_3 & \omega_4 & \omega_5\\
    \begin{block}{(cccccccccccccccc)}
     1 & 1 & 1 & 1 & 1  & 5 & 4 & 3 & 2  & 0 & 1 & 0 & -1 & 1 & -1 \\
    1 & 1 & 1 & 1 & 1  & 6 & 5 & 4 & 3  & 1 & 0 & -1 & 1 & -1 & 1 \\
    \end{block}
    \end{blockarray}
    \cdot
    \begin{blockarray}{c}
    \begin{block}{(c)}
    \hat{\theta_0}\\
    \hat{\theta}\\
    \end{block}
    \end{blockarray}
    \end{equation}.

\begin{figure}[H]
    \centering
    \begin{subfigure}[b]{0.32\textwidth}
        \centering
        \includegraphics[width=\textwidth]{img/lab_pred.PNG}
    \end{subfigure}
    \hfill
    \begin{subfigure}[b]{0.32\textwidth}
        \centering
        \includegraphics[width=\textwidth]{img/lab_pred.png}
    \end{subfigure}
    \hfill
    \vskip\baselineskip
    \begin{subfigure}[b]{0.44\textwidth}
        \centering
        \includegraphics[width=\textwidth]{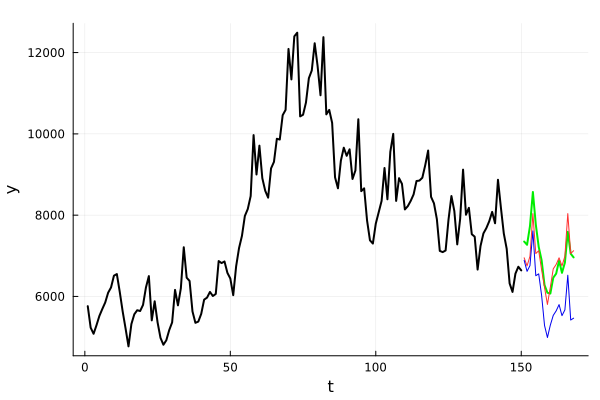}
    \end{subfigure}
    \hfill
    \begin{subfigure}[b]{0.44\textwidth}
        \centering
        \includegraphics[width=\textwidth]{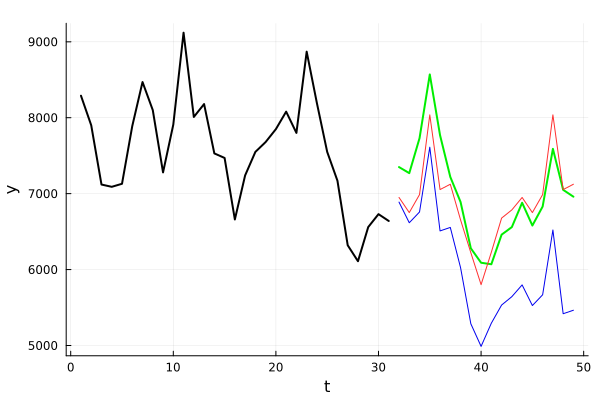}
    \end{subfigure}
    \hfill
    \vskip\baselineskip
    \begin{subfigure}[b]{0.42\textwidth}
        \includegraphics[width=\textwidth]{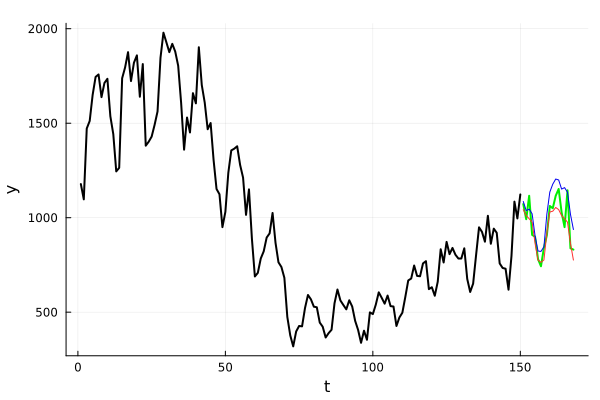}
    \end{subfigure}
    \hfill
    \begin{subfigure}[b]{0.42\textwidth}
        \includegraphics[width=\textwidth]{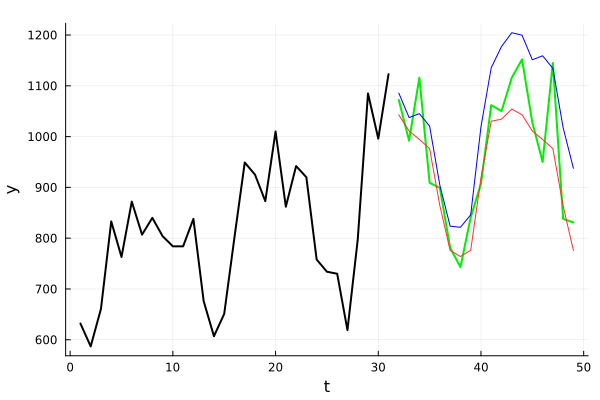}
    \end{subfigure}
    \hfill
    \vskip\baselineskip
    \begin{subfigure}[b]{0.42\textwidth}
        \includegraphics[width=\textwidth]{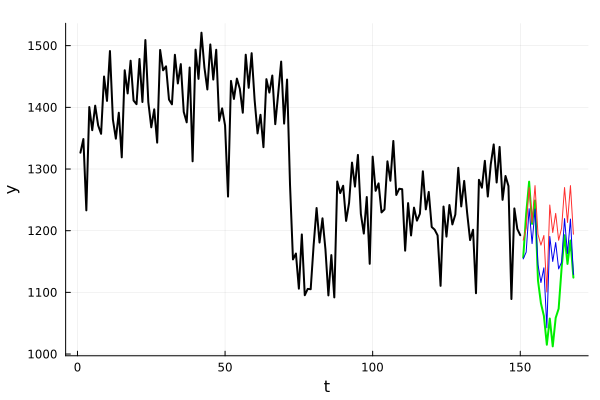}
        \caption{Forecast values of both the gaussian and the SSL $(\alpha=0.1-AIC)$ models in three different time series.}
    \end{subfigure}
    \hfill
    \begin{subfigure}[b]{0.42\textwidth}
        \includegraphics[width=\textwidth]{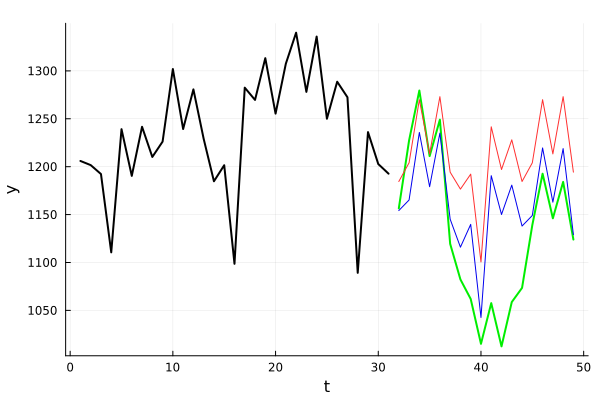}
        \caption{Zoom of the forecast values of both the gaussian and the SSL $(\alpha=0.1-AIC)$ models in three different time series from the M4 dataset.}
    \end{subfigure}
    \hfill
    \caption{Model Forecasts}
    \label{forecasts}
\end{figure}

We now present in Figure \ref{forecasts} an illustrative example of our proposed methodology's forecasting capabilities and compared it with its Gaussian counterpart across three time series sourced from the M4 dataset. 

It is noteworthy that both the Gaussian benchmark and the SSL model illustrate proficiency in capturing level, slope, and seasonal shifts, thereby furnishing credible forecasts. A more comprehensive evaluation of the forecasting performance of SSL is presented in Section \ref{pr}.

\section{State Space Learning: Refinements and Extensions}

In this section, we explore several extensions and refinements to enhance the capabilities of the state space learning framework. These extensions broaden the applicability and robustness of the framework across various domains and analytical challenges. Firstly, we introduce a subset selection method tailored to accommodate exogenous variables within high-dimensional spaces, enabling more comprehensive modeling of time series in a big-data world. Secondly, we delve into outlier detection and treatment as well as techniques to avoid overfitting, a  crucial development for ensuring the reliability and accuracy of model outputs in the presence of noisy or erroneous data points. Lastly, we discuss the extension of the framework beyond traditional time series analysis, allowing for its utilization in many engineering applications that resort to linear state space formulation. 

\subsection{Subset Selection of Exogenous Variables}

It is worth noting that the proposed methodology and its associated estimation procedure possess the capability to seamlessly adapt to best subset selection with a slight modification in its formulation
\begin{align}
     y_t =& \ \mu_1 + (t-1)\nu_1 + \sum_{\tau=2}^{t-1}(t-\tau)\zeta_\tau + \sum_{\tau=2}^{t}\xi_\tau \nonumber\\
     &\gamma_{m_t} +\sum_{\tau \in M_t}(\omega_{\tau} -  \omega_{\tau-1}) +  X_t\beta + \varepsilon_t .
\end{align}

In this adapted framework, a new component $X_t \beta$ emerges, where $X_t$ contains the observations of the exogenous features, and $\beta$ represents the coefficients associated with these exogenous variables, subject to estimation by the model. 

Note that the adapted estimation procedure will be modified. Initially, in the first step (Equation \ref{adalasso_equation1}), we redefine $\theta = (\xi_2,\ldots,\xi_T, \zeta_2,\ldots,\zeta_T, \omega_s, \ldots,\omega_T, \beta)$. In the second step (Equation \ref{adalasso_equation2}) there will be the addition of the term $\sum_{i=1}^p w_{\beta_i} \left((1-\alpha)\frac{1}{2}    \|\beta_i\|_2^2 + \alpha    \|\beta_i\|_1 \right)$, where $w_{\beta_i}=|\hat\beta_i|^{-1}$ and $p$ is the number of exogenous features, $\hat\beta_i$ is the estimated value of $\beta_i$ on the first step of the estimation procedure.

Furthermore, the integration of this term into the forecasting framework is straightforward. To accomplish this, we adapt Equation \ref{for_eq}:
\begin{align}
\begin{split}
\hat{y}_{t} = \mu_1 + (t-1)\nu_1 + & \sum_{\tau=2}^{T-1}(t-\tau)\zeta_\tau + \sum_{\tau=2}^{T}\xi_\tau + \gamma_{m_t}  +\sum_{\tau \in M_{T}}(\omega_{{\tau}} -  \omega_{\tau-1}) + \hat{X}_t \beta, \quad \forall t > T
\end{split},
\end{align}\label{for_eqex}wherein $\hat X_t$ denotes the predicted values of the exogenous variables for $t>T$.

 By integrating a distinctive framework, the proposed methodology not only concurrently selects and estimates intrinsic time series components but also fills a critical void in the literature by simultaneously addressing the selection and estimation of exogenous features in both stationary and non stationary time series. This enhancement significantly augments the model's versatility and utility in statistical modeling and forecasting applications.

Note that the forecasting via extrapolation procedure can be utilized in the presence of exogenous features as long as a forecast for each variable is provided for the underlying model in order to build their columns in the forecast regression matrix. This leads to an adaptation of Equation \ref{for_eq} with the addition of the term $\hat X_t$ (the forecast of the exogenous variables at a given time ``t").

\subsection{Robust Forecasting: Handling Outliers, Identification and Overfitting}\label{sec:robust_forecast}

In this section we will propose a methodology to handle outlier observations, an identification issue for the extrapolation procedure and a strategy to handle potential overfitting. 

With the regression formulation we are able to adapt our model to account for the presence of sporadic values that deviate from the natural behavior of the model and must be disregarded in estimation and forecasting or simulation. These observations, commonly known as outliers, require a distinct treatment to avoid misleading results. To address this, we introduce additional columns of dummy variables to the high-dimensional regression matrix, equivalent to adding an identity matrix to the underlying model. This enables regularization techniques to sparsely identify and isolate outliers while simultaneously estimating the best-fit model for the studied time series. By incorporating this approach, we can enhance the accuracy and reliability of our predictions and simulations, particularly in situations where outliers are prevalent.

The formulation containing the outlier component $D_{\tau,t} \,o_t$ is presented below:

\begin{align}
     y_t =&  \mu_1  + \sum_{\tau=2}^{t}\xi_\tau + (t-1)\nu_1 + \sum_{\tau=2}^{t-1}(t-\tau)\zeta_\tau + \gamma_{m_{t}} + \sum_{\tau \in M_{t}}(\omega_{{\tau}} -  \omega_{\tau-1}) +  X_t\beta + \sum_{\tau=1}^tD_{\tau,t} \,o_\tau + \varepsilon_t,
\end{align}
where $D_{\tau,t}$ is a dummy term that assumes value equal to 1 when $\tau=t$ and 0 otherwise. Moreover, $o_\tau$ represent the coefficient estimated for each outlier component. Note that this procedure is similar to the one proposed in \cite{boyd} in the L1 filtering context.

Analogous to the exogenous features, the adapted estimation procedure will be modified. In Equation \ref{adalasso_equation1}), we redefine  $\theta = (\xi_2,\ldots,\xi_T, \zeta_2,\ldots,\zeta_T, \omega_s, \ldots,\omega_T, \beta, o_1,\ldots,o_T)$. In Equation \ref{adalasso_equation2}, we add the term $\sum_{t=1}^T w_{o_t} \left((1-\alpha)\frac{1}{2}    \|o_t\|_2^2 + \alpha    \|o_t\|_1 \right)$, where $w_{o_t}=|\hat o_t|^{-1}$ and $\hat o_t$ is the estimated value of $o_t$ on the first step of the estimation procedure.

Note that the addition of the outlier component in the forecasting procedure is a straightforward process. Specifically, the outlier component is anticipated not to affect the forecast due to the constant 0 value of the dummy term $D_{\tau,t}$ for $t > T$. Therefore, the forecasting equation \ref{for_eq} will remain the same.

After handling outliers, we propose small modifications in the estimation to robustify the forecasting procedure. First, it is crucial to recognize that the regression matrix underlying the proposed model, as illustrated in Equation \ref{Eq:matrix1}, encounters an identification challenge due to the indistinguishable nature of the regressors associated with terms ($\xi_T, \zeta_{T-1}, \omega_T, o_T$). Note that this distinction has no impact over the estimation procedure. However, for the extrapolation procedure it is extremely relevant as it it will state what time series dynamic was changed at time T. As it is not possible to affirm that a change over a level, slope or seasonal component occurred with a single observation ``$y_T$" we opted to constrain the values of $``\xi_T", ``\zeta_{T-1}" and ``\omega_T"$ at 0 and keep $o_T$ unconstrained to robustify the forecast procedure.

More generally, the final terms of the innovation components are estimated based on a limited set of observations, leading to overfitting the model to the most recent data. This is more significant for the slope innovation $\zeta_t$, where spurious slope changes at the end of a time series can degrade forecasting accuracy. We argue that such spurious changes might result from the model using too few observations to infer a slope or seasonal shock, which is not a parsimonious approach. To address this challenge, we propose constraining the latest terms of the $\zeta_t$ and $\omega_t$ shocks to zero for $t > T-s$. In our numerical experiments, we implement a threshold with $s=12$ (given that the tested time series have a monthly granularity) for these components. Specifically, we apply the constraints $\zeta_\tau = 0 \; \forall \tau \in (T-12, T]$ and $\omega_\tau = 0 \; \forall \tau \in (T-12, T]$.

\subsection{SSL as an initialization procedure for for the Gaussian Basic Structural Model}

As discussed in Section \ref{Introduction}, the traditional state-space model framework depends on good initialization points, given that its estimation procedure relies on nonlinear optimization. 

Another use case of the proposed State Space Learning framework is that it leads to a straightforward initialization for its Gaussian counterpart. For the Basic Structural Model case, the initialization values (the variances of innovation terms and residuals) can be obtained as $\hat \sigma_\kappa^2 = \|\kappa\|_2^2, \forall \kappa \in \{\xi,\zeta,\omega, \varepsilon\}$.

It is noteworthy that the traditional Basic Structural Model Explanatory, which incorporates exogenous features, lacks the capability to efficiently perform best subset selection. This limitation is particularly significant in the context of today's big data landscape. In contrast, the SSL framework addresses this shortcoming by incorporating an efficient best subset selection process. 

Thus, the SSL results can be utilized to reduce the dimensionality of the set of explanatory variables for the Basic Structural Model. In this case, the Basic Structural Model can consider only the pre-selected variables, whose coefficients ($\beta$) estimated within the SSL methodology are non-zero. Subsequently, all parameters of the Basic Structural Model can be initialized using the values derived from the SSL estimation methodology.

Expanding the initialization process to accommodate the outlier component detailed in Section \ref{sec:robust_forecast} involves a strategic approach. Specifically, observations ($y_\tau$) characterized by non-zero outlier components ($o_\tau$) are treated as missing values. Leveraging the flexibility of traditional state-space models in handling missing data, this treatment effectively excludes outlier-influenced observations during the initialization phase. By doing so, the proposed initialization process for the Basic Structural Model based on the SSL estimates mitigates the impact of outlier observations, thereby enhancing the robustness of subsequent model estimation and analysis.

\subsection{Generalized State-Space Learning}\label{sec:general_ssl}

A general additive state-space model can be written as:
\begin{align}
    y_t & = Z_t \alpha_t + d_t + \varepsilon_t \label{gen_1}\\
     \alpha_{t} & = T_{t-1} \alpha_{t-1} + c_{t-1} + R_{t-1} \eta_{t-1}, \label{gen_2}
\end{align}
where $y_t$ is typically an observed data point in a time series analysis. The term $Z_t$ represents the observation matrix at time t, which maps the state variable $\alpha_t$ to the observed variable $y_t$. The state variable $\alpha_t$ captures the underlying, unobserved components that evolves over time. The terms $d_t$ and $\varepsilon_t$ represent the deterministic and random components, respectively, of the observation equation. $\alpha_{t}$ represents the state variable at the next time step, which evolves over time. The term $T_t$ denotes the state transition matrix, indicating how the state variable at time t evolves to the state variable at time t+1. The terms $c_t$ represent the deterministic component of the state equation. $R_t$ denotes the transition matrix for the random component or process noise and $\eta_t$ represents the unobserved, random disturbances or shocks that affect the evolution of the state variable $\alpha_t$.

We now extend the proposition that a basic structural model can be written as a high dimension regression to any additive state-space model. This overarching generalization holds significant implications, given its potential applicability across various disciplines reliant on state-space models. It introduces an alternative estimation approach for the Kalman Filter, thereby expanding the methodological repertoire available to researchers and practitioners in these fields. 

To facilitate the following proof, we start by iterating the state equation as: 

\begin{align}
    \alpha_{t} &= T_{t-1} \alpha_{t-1} + c_{t-1} + R_{t-1}\eta_{t-1}\\ 
    \alpha_{t} &= T_{t-1} \left(T_{t-2} \alpha_{t-2} + c_{t-2} + R_{t-2}\eta_{t-2}\right) + c_{t-1} + R_{t-1}\eta_{t-1}\\ 
    ...\\
    \alpha_{t} &= \sum_{j=1}^{t-1}T_{t-j}\alpha_1 + \sum_{j=1}^{t-1}c_{t-j}\prod_{i=1}^{t-j}T_{t-i}+\sum_{j=1}^{t-1}R_{t-j}\eta_{t-j}\prod_{i=1}^{t-j}T_{t-i}.\label{state_eq_gen}
\end{align}

\begin{proposition}\label{prop_gen}
Any state-space model of the form of Equations \ref{gen_1}-\ref{gen_2} can be written as a high dimension regression: $y_t = Z_t \left(\sum_{j=1}^{t-1}T_{t-j}\alpha_1 + \sum_{j=1}^{t-1}c_{t-j}\prod_{i=1}^{t-j}T_{t-i}+ \sum_{j=1}^{t-1}R_{t-j}\eta_{t-j}\prod_{i=1}^{t-j}T_{t-i}\right) + d_t + \varepsilon_t$. 
\end{proposition}

\begin{proof}
We also prove Proposition \ref{prop_gen} by induction of the state equation. Our base case $\alpha_2 = T_1\alpha_1+c_1+R_1\eta_1$ holds true for $t=2$. 

The induction hypothesis $y_t = Z_t \left(\sum_{j=1}^{t-1}T_{t-j}\alpha_1 + \sum_{j=1}^{t-1}c_{t-j}\prod_{i=1}^{t-j}T_{t-i}+ \sum_{j=1}^{t-1}R_{t-j}\eta_{t-j}\prod_{i=1}^{t-j}T_{t-i}\right) + d_t + \varepsilon_t$ (for $t>2$) can be confirmed by substituting Equation \ref{state_eq_gen} in the observation Equation \ref{gen_1}.

\end{proof}

Note that as the system variables ($Z_t$, $T_t$, $d_t$, $c_t$, $R_t$) are known. Therefore we can create a new y' variable as: 
\begin{align}
    y'_t = y_t - Z_t\sum_{j=1}^{t-1}c_{t-j}\prod_{i=1}^{t-j}T_{t-i} - d_t,
\end{align}
and finally, we are able to present the general representation of a state-space formulation as a high dimension regression:
\begin{align}
    y'_t = Z_t \sum_{j=1}^{t-1}T_{t-j}\alpha_1 + Z_t\sum_{j=1}^{t-1}R_{t-j}\eta_{t-j}\prod_{i=1}^{t-j}T_{t-i} + \varepsilon_t.
\end{align}

This equation can be divided in two terms. The first one ($Z_t \sum_{j=1}^{t-1}T_{t-j}\alpha_1$) is referent to the estimation of the initial states $\alpha_1$ (referent to the $\theta_0$ variable described in Section \ref{estimation_p}).  The second term ($Z_t\sum_{j=1}^{t-1}R_{t-j}\eta_{t-j}\prod_{i=1}^{t-j}T_{t-i}$) is referent to the estimation noise terms $\eta$ (referent to the $\theta$ variable described in Section \ref{estimation_p}). The terms $Z_t$, $R_t$ and $T_t$ are known as system matrices and will form the a set of columns in the regression matrix.

Finally, we could employ the exactly proposed estimation procedure \ref{estimation_p} by grouping the state variables weights and not penalizing an initial state equivalent to an intercept (if existing).

Analogous to traditional state-space models, this versatile framework extends its scope beyond time series analysis, finding application in various domains. Analogous to its Gaussian counterpart, this generalized model may be object of study in vehicle tracking, physics, finance, and more. Additionally, it is instrumental in machine learning, speech recognition, natural language processing, and anomaly detection tasks, enabling the modeling of latent factors. These diverse applications highlight the wide-ranging significance of this generalized structural model, enhancing analytical capabilities across multidisciplinary domains.

\section{Numerical Results}

In this section, we conduct a comprehensive numerical assessment of the capabilities of the State Space Learning (SSL) framework in the context of time series analysis. Initially, we demonstrate the predictive prowess of SSL through extensive empirical analysis, evaluating its out-of-sample performance across 48,000 monthly time series from the M4 dataset. Through this experiment, we rigorously compare SSL against selected benchmarks, employing various model selection criteria and robust forecasting strategies to address outliers and overfitting issues. Our results underscore the superior performance of SSL over these benchmarks. Furthermore, we present simulation results illustrating SSL's effectiveness in addressing the best subset selection problem, particularly in selecting and estimating the impact of exogenous variables within high-dimensional spaces. Our findings provide compelling evidence of SSL's superior performance in best subset selection when compared to selected benchmarks.

\subsection{Empirical Evaluation of SSL Predictive Power}\label{pr}
In this section, we undertake a thorough evaluation of the SSL robust forecasting methodology. We compare the performance of two variants of the SSL model, namely SSL and SSL-O (which handles outliers), against the benchmark Gaussian Basic Structural Model, the SSL-initialized Gaussian Basic Structural Model, and the auto Sarima benchmark \cite{sarima} using the M4 dataset.

The M4 dataset comprises a vast collection of 100,000 time series spanning various frequencies and originating from diverse sectors, including business, finance, and economic forecasting. For our analysis, we focus on a subset of 48,000 monthly time series from this dataset. Notably, the forecasting horizon for these monthly time series extends to 18 steps ahead.

To rigorously assess our results, we utilize standard performance metrics including sMAPE, MASE, and OWA, consistent with those adopted in \cite{m4competitorsguide}. The sMAPE measures error scaling using the average between forecasted and actual values. MASE scales error by the average error of a naive predictor, which replicates observations measured s periods in the past, thereby accounting for seasonality. Additionally, we employ a competition-specific metric, Overall Weighted Average (OWA), used to rank competition entries in the M4 competition \cite{m4competitorsguide}. OWA is designed such that a seasonally-adjusted naive forecast (Naïve 2 \cite{naive2}) achieves a score of 1.0. Specifically, the OWA of sMAPE and MASE is computed by dividing their total value by the corresponding value of Naïve 2 to obtain the relative sMAPE and relative MASE, respectively, and then calculating their simple arithmetic mean.

For a given forecast horizon H, a T length time series $y$, we define
\begin{align}
sMAPE&=\frac{2}{H} \sum_{t=1}^{H}\frac{|y_{T+1}-\hat{y}_{T+t}|}{|y_{T+t}|+|\hat{y}_{T+t}|}\cdot 100(\%)\\
MASE&= \frac{1}{H}\frac{\sum_{t=T+1}^{T+H}|y_t-\hat{y}_t|}{\frac{1}{T-s}\sum_{t=s+1}^T|y_t-y_{t-s}|}\\
OWA&=\frac{1}{2}\left(\frac{sMAPE}{sMAPE^{NAIVE2}} + \frac{MASE}{MASE^{NAIVE2}}\right),
\end{align} where we denote the time series forecast as $\hat{y}$ and s as the periodicity of the data (12 for monthly series).

To compare the SSL against the benchmark models, we utilized the robust forecasting technique presented in Section \ref{sec:robust_forecast} with different values of alpha (the coefficient that sets the convex combination between Lasso and Ridge Regression) and different information criteria to obtain the $\lambda$ value. We also utilized the result of the Gaussian Basic Structural Model with the initialization of the best-tested SSL model in the OWA metric ($\alpha$=0.1 and AIC as information criteria). The presented results were obtained using Julia programming language with the GLMNet.jl \citep{glmnetjl} package to estimate elastic net models and the python package \citep{seabold2010statsmodels} to estimate the Gaussian Basic Structural Model. For Both SSL and SSL-O we calibrated the number of train observations to the last five years of data (60 observations) as this leads to slightly better results for this particular model, which indicates that it does not need much data to present robust forecasts. We also compare the results against the benchmark results of Auto Sarima (published in the M4 competition results \citep{m4competitorsguide}). 

Moreover, we present the results of the applied model in two different ways: one considering the outlier component --- SSL-O --- and another one without it (SSL). We present Table \ref{my-label} comparing the mean in each metric for the benchmark models and the best performing configurations for the proposed methodology.

\begin{table}[H]
\centering
\caption{Performance on the monthly time series of M4 dataset (lower values are better).}
\label{my-label}
\begin{tabular}{|c|c|c|c|}
    \hline
        \textbf{Model} & \textbf{MASE} & \textbf{sMAPE} & \textbf{OWA} \\
\hline\hline
        Gaussian Basic Structural Model & 1.012 & 15.689 & 1.020 \\ \hline
        SSL initialized Gaussian Basic Structural Model & 0.965 & 14.894 & 0.970 \\ \hline
        AUTO SARIMA &  \textbf{0.930} & 13.443 & 0.903 \\ \hline
        SSL-O (AIC, $\alpha$ = 0.0) & 0.984 & 13.942 & 0.946 \\ \hline
        SSL (AIC, $\alpha$ = 0.0) & 0.986 & 14.367 & 0.961 \\ \hline
        SSL-O (AIC, $\alpha$ = 0.1) & 0.936 & \textbf{12.975} & \textbf{0.89} \\ \hline
        SSL (AIC, $\alpha$ = 0.1) & 0.992 & 14.208 & 0.959 \\ \hline
        SSL-O (AIC, $\alpha$ = 0.3) & 0.954 & 13.103 & 0.903 \\ \hline
        SSL (AIC, $\alpha$ = 0.3) & 1.004 & 14.252 & 0.966 \\ \hline
        SSL-O (AIC, $\alpha$ = 0.5) & 0.962 & 13.168 & 0.909 \\ \hline
        SSL (AIC, $\alpha$ = 0.5) & 1.008 & 14.25 & 0.968 \\ \hline
        SSL-O (AIC, $\alpha$ = 0.7) & 0.965 & 13.195 & 0.911 \\ \hline
        SSL (AIC, $\alpha$ = 0.7) & 1.009 & 14.258 & 0.969 \\ \hline
        SSL-O (AIC, $\alpha$ = 0.9) & 0.966 & 13.214 & 0.912 \\ \hline
        SSL (AIC, $\alpha$ = 0.9) & 1.01 & 14.269 & 0.969 \\ \hline
        SSL-O (AIC, $\alpha$ = 1.0) & 0.966 & 13.223 & 0.912 \\ \hline
        SSL (AIC, $\alpha$ = 1.0) & 1.01 & 14.272 & 0.97 \\ \hline
        SSL-O (BIC, $\alpha$ = 0.0) & 0.984 & 13.942 & 0.946 \\ \hline
        SSL (BIC, $\alpha$ = 0.0) & 0.986 & 14.367 & 0.961 \\ \hline
        SSL-O (BIC, $\alpha$ = 0.1) & 1.183 & 14.654 & 1.064 \\ \hline
        SSL (BIC, $\alpha$ = 0.1) & 1.007 & 14.158 & 0.964 \\ \hline
        SSL-O (BIC, $\alpha$ = 0.3) & 0.954 & 13.106 & 0.903 \\ \hline
        SSL (BIC, $\alpha$ = 0.3) & 1.003 & 14.244 & 0.965 \\ \hline
        SSL-O (BIC, $\alpha$ = 0.5) & 0.962 & 13.168 & 0.909 \\ \hline
        SSL (BIC, $\alpha$ = 0.5) & 1.007 & 14.244 & 0.967 \\ \hline
        SSL-O (BIC, $\alpha$ = 0.7) & 0.965 & 13.195 & 0.911 \\ \hline
        SSL (BIC, $\alpha$ = 0.7) & 1.008 & 14.255 & 0.968 \\ \hline
        SSL-O (BIC, $\alpha$ = 0.9) & 0.966 & 13.214 & 0.912 \\ \hline
        SSL (BIC, $\alpha$ = 0.9) & 1.009 & 14.265 & 0.969 \\ \hline
        SSL-O (BIC, $\alpha$ = 1.0) & 0.966 & 13.223 & 0.913 \\ \hline
        SSL (BIC, $\alpha$ = 1.0) & 1.009 & 14.268 & 0.969 \\ \hline
    \end{tabular}
\end{table}

A noteworthy finding emerges from the analysis, revealing the notable superiority of both the top-performing SSL and SSL-O models in terms of all metrics against its Gaussian Counterpart. These models exhibited remarkable outcomes, surpassing not only the benchmark Basic Structural Model but also against its initialized version. Intriguingly, the initialized Gaussian model demonstrated superior performance across all metrics when compared to the non initialized counterpart. Moreover, the proposed methodology got better results than the auto Sarima benchmark in terms of two of the studied metrics including the official metric of the M4 competition (OWA). 

Further, the analysis indicates that the outlier component played a crucial role in enhancing the model's efficacy, as all of the variants of the proposed formulation outperformed their counterparts across all metrics.

Moreover,  the performance improvement observed with the initialized Gaussian model, as opposed to its non-initialized counterpart, is particularly significant. These results substantiate the assertion that the SSL model serves as an effective initializer for the conventional Basic Structural Model.

Taken together, these outcomes showcase the potential of the SSL model to outshine its benchmark counterpart in forecasting tasks. The efficacy of this approach highlights its significance in advancing the field of predictive modeling and promoting better decision-making in various domains.

\subsection{Best Subset Selection: A Simulation Assessment}

In the realm of conventional time series modeling, consensus on the optimal subset selection method remains elusive. While regularization techniques like Lasso offer a promising avenue, they rely on stationary hypotheses that often fail to hold in many real-world applications. Conversely, stepwise approaches such as forward and backward selection, though heuristic, come with significant computational costs.

In this study, we embark on evaluating the efficacy of our proposed methodology for best subset selection. We benchmark our approach against two established methods outlined below:

\begin{itemize}
    \item SS+FW: A stepwise forward selection (FW) process, where exogenous features are iteratively added to the state-space (SS) Gaussian basic structural model until a predetermined information criterion ceases to improve. This method utilizes a basic structural model estimated with \cite{seabold2010statsmodels}.
    \item SS+AdaLasso: A two-stage procedure involving the estimation of a traditional state space (SS) Gaussian basic structural model (via \cite{seabold2010statsmodels}), followed by the removal of trend and seasonal components. Subsequently, an Adaptive Lasso (AdaLasso) is performed on the remaining values to select the optimal subset (we utilize $\alpha=1$, where $\alpha$ is the parameter that balances the L2 and L1 penalties).
\end{itemize}

To commence our simulation experiment, we first outline the data generation process. Our approach begins by constructing the dependent variables, which are derived from time series generated using Basic Structural Models:

\begin{align}
    y_t &= \mu_t + \gamma_t + X_t\beta+ \varepsilon_t, &\quad \varepsilon \sim N(0, \sigma_\varepsilon^2) \label{eq1_2}\\
    \mu_{t} &= \mu_{t-1} + \nu_{t} + \xi_t, &\quad \xi \sim N(0, \sigma_\xi^2) \label{eq2_2}\\
    \nu_{t} &= \nu_{t-1} + \zeta_t, &\quad \zeta \sim N(0, \sigma_\zeta^2)\label{eq3_2}\\
    \gamma_{t} &= -\sum_{j=1}^{s-1} \gamma_{t - j} + \omega_t, &\quad \omega \sim N(0, \sigma_\omega^2)\label{eq4_2},
\end{align}
where for this experiment we define $\mu_0= 1$, $\nu_0 = 0.001$, $\gamma_0 = $[1.5, 2.6, 3.0, 2.6, 1.5, 0.0, -1.5, -2.6, -3.0, -2.6, -1.5], $\sigma_\xi = 0.5$, $\sigma_\zeta = 0.001$, $\sigma_\omega = 0.3$ and , and $\sigma_\varepsilon = 0.2$.

Moreover, we defined each exogenous vector $X_t$ as a Seasonal Autoregressive Integrated Moving Average (SARIMA) process with parameters $(p, 0, q)(P, 0, Q)_{12}$. Within this process, each coefficient was stochastically drawn from a uniform distribution $\theta, \Theta, \phi, \Phi \sim N(0, 0.2)$. Furthermore, $p$ and $q$ were selected from $\{0, 1, 2, 3, 4, 5\}$, while $P$ and $Q$ were chosen from $\{0, 1\}$. Additionally, an error term, sampled from a normal distribution $N(0, 3)$, was introduced for each exogenous feature. The coefficients $\beta$ of the relevant features were assigned as 1.

To conduct a thorough analysis, we explored a range of combinations denoted by the tuple $(M, K)$, where M is the number of candidates and K is the number of true relevant features. Specifically, we varied $M$ within the set $\{3, 5, 8, 10\}$ and $K$ within $\{50, 100\}$. For each configuration of $(M, K)$, we performed a Monte Carlo simulation comprising 500 samples. 

An essential consideration arises upon parameter sorting in the described SARIMA framework: the resulting process may exhibit non-stationarity and lack invertibility. To ensure equitable comparisons with the proposed benchmarks, which lack the capability to manage non-stationary exogenous variables, we implement a filtering mechanism. This involves restricting exogenous variables to only include those generated within the SARIMA framework that successfully pass the unit root test \cite{brockwell1991time} for both Autoregressive and Moving Average terms. 

We train the SSL model for a fixed value of $\alpha = 1$ (for a fair comparison with SS+AdaLasso benchmark)  and test two information criteria (AIC and BIC) for selection of the $\lambda$ parameter.

For the first benchmark model, we employed a forward selection model to identify significant predictors within a basic structural explanatory (estimated via \cite{seabold2010statsmodels}) framework. Forward selection is a feature selection method commonly used in statistical modeling to iteratively add variables to the model, starting with zero variables. At each step, the variable that provides the greatest improvement in a given information criteria is added, and the process continues until the stopping criterion is met. In our case, we tested both the AIC and BIC as stopping criteria.

For our second benchmark model, we begin by estimating a basic structural model using the methodology described in \cite{seabold2010statsmodels}. Subsequently, we remove the level and seasonal components from the original time series $y_t$. This process yields a residual series defined as
$ y^*_t = y_t - \hat{\mu}_t - \hat{\gamma}_t, $ where $\hat\mu_t$ and $\hat\gamma_t$ are the level and seasonal component obtained with a Gaussian structural model.
We then apply an Adaptive Lasso technique to $y^*_t$, facilitating best subset selection. This two-step approach combines state-of-the-art methods for estimating time series basic structural models and performing variable selection through statistical learning.

In contrast, our proposed SSL approach integrates these two steps into a single cohesive process. By co-optimizing both component extraction and exogenous variable selection, SSL streamlines the modeling procedure and potentially enhances performance compared to the separate two-step process.
Finally, we run our controlled test and present the results Tables \ref{TableSim1}-\ref{TableSim2}.

Table \ref{TableSim1} depicts the variable selection outcomes for each model selection technique. Various pertinent statistics are presented: Panel (a) showcases the proportion of repetitions where the accurate model was selected, indicating inclusion of all relevant variables and exclusion of all irrelevant regressors from the final model; Panel (b) illustrates the proportion of repetitions where all relevant variables were included; Panel (c) exhibits the proportion of relevant variables that were incorporated; Panel (d) illustrates the proportion of irrelevant variables that were excluded; Panel (e) displays the mean count of variables included; Panel (f) showcases the average count of irrelevant variables included.

\begin{table}[H]
\footnotesize
\begin{center}
\begin{tabular}{|c|cc|cc|cc|cc|cc|cc|}
\hline
\multicolumn{1}{c}{} & \multicolumn{2}{c}{SSL (AIC)} & \multicolumn{2}{c}{SSL (BIC)} & \multicolumn{2}{c}{SS+AdaLasso (AIC)} & \multicolumn{2}{c}{SS+AdaLasso (BIC)} & \multicolumn{2}{c}{SS+FW (AIC)} & \multicolumn{2}{c}{SS+FW (BIC)} \\ \hline
\multicolumn{13}{c}{}                                                                      \\
M\textbackslash{}K   & 50                      & 100                  &                       50                      & 100                  &                       50                 & 100            &                       50                 & 100    
&                       50                 & 100  
&                       50                 & 100  
\\ \hline
\multicolumn{13}{|c|}{Panel (a): Correct Sparsity Pattern} \\ \hline
3 & 0.652 & 0.552 & 0.76 & 0.682 & 0.002 & 0.002 & 0.178 & 0.018 & 0.488 & 0.458 & 0.492 & 0.458 \\ 
5 & 0.714 & 0.634 & 0.778 & 0.718 & 0.0 & 0.0 & 0.064 & 0.002 & 0.134 & 0.132 & 0.132 & 0.13 \\ 
8 & 0.79 & 0.664 & 0.856 & 0.734 & 0.0 & 0.0 & 0.002 & 0.0 & 0.004 & 0.008 & 0.004 & 0.002 \\ 
10 & 0.796 & 0.628 & 0.844 & 0.674 & 0.0 & 0.0 & 0.0 & 0.0 & 0.002 & 0.002 & 0.0 & 0.002 \\ \hline
\multicolumn{13}{|c|}{Panel (b): True Model Included}                    \\ \hline
3 & 0.996 & 0.994 & 0.996 & 0.994 & 0.994 & 0.988 & 0.99 & 0.986 & 0.654 & 0.63 & 0.656 & 0.632 \\ 
5 & 0.99 & 0.986 & 0.99 & 0.986 & 0.964 & 0.934 & 0.916 & 0.93 & 0.14 & 0.14 & 0.138 & 0.138 \\ 
8 & 0.984 & 0.986 & 0.984 & 0.986 & 0.878 & 0.824 & 0.762 & 0.822 & 0.004 & 0.008 & 0.004 & 0.002 \\ 
10 &  0.984 & 0.942 & 0.984 & 0.942 & 0.824 & 0.574 & 0.662 & 0.57 & 0.002 & 0.002 & 0.0 & 0.002 \\ \hline
\multicolumn{13}{|c|}{Panel (c): Fraction of Relevant Variables Included}                                     \\ \hline
3 & 0.997 & 0.995 & 0.997 & 0.995 & 0.996 & 0.994 & 0.994 & 0.993 & 0.895 & 0.883 & 0.895 & 0.883 \\ 
5 & 0.991 & 0.99 & 0.991 & 0.99 & 0.982 & 0.975 & 0.96 & 0.973 & 0.666 & 0.668 & 0.665 & 0.665 \\ 
8 &  0.988 & 0.99 & 0.988 & 0.99 & 0.96 & 0.962 & 0.897 & 0.962 & 0.462 & 0.429 & 0.439 & 0.408 \\ 
10 & 0.989 & 0.972 & 0.989 & 0.972 & 0.965 & 0.911 & 0.856 & 0.906 & 0.377 & 0.321 & 0.338 & 0.29 \\ \hline
\multicolumn{13}{|c|}{Panel (d): Fraction of Irrelevant Variables Excluded}                                                  \\ \hline
3 & 0.986 & 0.989 & 0.991 & 0.992 & 0.231 & 0.457 & 0.911 & 0.516 & 0.996 & 0.998 & 0.996 & 0.998 \\ 
5 & 0.99 & 0.991 & 0.993 & 0.994 & 0.189 & 0.462 & 0.83 & 0.479 & 0.999 & 0.999 & 0.999 & 0.999 \\ 
8 & 0.992 & 0.992 & 0.995 & 0.994 & 0.184 & 0.462 & 0.711 & 0.467 & 0.996 & 0.997 & 0.997 & 0.997 \\ 
10 & 0.993 & 0.988 & 0.995 & 0.99 & 0.162 & 0.473 & 0.666 & 0.481 & 0.994 & 0.995 & 0.995 & 0.995 \\ \hline
\multicolumn{13}{|c|}{Panel (e): Number of Included Variables}                                                             \\ \hline
3 &  3.646 & 4.058 & 3.398 & 3.72 & 39.13 & 55.626 & 7.15 & 49.966 & 2.89 & 2.866 & 2.86 & 2.84 \\ 
5 & 5.402 & 5.806 & 5.274 & 5.53 & 41.388 & 55.992 & 12.47 & 54.312 & 3.364 & 3.398 & 3.354 & 3.38 \\ 
8 & 8.222 & 8.674 & 8.104 & 8.456 & 41.968 & 57.22 & 19.31 & 56.74 & 3.866 & 3.744 & 3.66 & 3.55 \\ 
10 & 10.178 & 10.784 & 10.092 & 10.618 & 43.154 & 56.516 & 21.91 & 55.81 & 3.998 & 3.7 & 3.58 & 3.316 \\ \hline
\multicolumn{13}{|c|}{Panel (f): Number of Included Irrelevant Variables} \\ \hline
3 & 0.654 & 1.074 & 0.406 & 0.736 & 36.142 & 52.644 & 4.168 & 46.988 & 0.204 & 0.216 & 0.176 & 0.192 \\ 
5 & 0.446 & 0.858 & 0.318 & 0.582 & 36.48 & 51.116 & 7.67 & 49.448 & 0.032 & 0.06 & 0.028 & 0.054 \\ 
8 & 0.318 & 0.752 & 0.2 & 0.534 & 34.286 & 49.524 & 12.132 & 49.048 & 0.166 & 0.308 & 0.144 & 0.282 \\ 
10 & 0.286 & 1.06 & 0.2 & 0.898 & 33.502 & 47.404 & 13.348 & 46.746 & 0.23 & 0.486 & 0.196 & 0.416 \\ \hline
    \end{tabular}
    \caption{Statistics concerning model selection for each sample configuration. Panel (a), the fraction of instances where the accurate model was chosen is depicted. Panel (b) showcases the proportion of repetitions in which all pertinent variables were incorporated. Panel (c) illustrates the fraction of relevant variables that were included in the selected models. Panel (d) exhibits the ratio of irrelevant variables that were excluded from the chosen models. Panel (e) displays the mean count of variables incorporated across all selected models. In Panel (f), the average number of irrelevant variables included in the chosen models is depicted. Here, ``p" denotes the total number of candidate variables, while ``q" represents the quantity of relevant variables in consideration.}
\label{TableSim1}
\end{center}
\end{table}

It is noteworthy that the State Space Learning framework consistently surpassed the benchmarks across all examined scenarios (in terms of Correct Sparsity Pattern), particularly demonstrating remarkable advancements in instances characterized by elevated counts of candidate variables and true features. The State Space Learning framework was able to hold and even increase the ration of correct sparsity pattern in more complex scenarios. It is interesting that both benchmarks got better results when the number of true relevant variables was low (3), but got increasingly bad results as this number increases.

Furthermore, our proposed framework consistently exhibited a parsimonious selection of variables, particularly utilizing BIC as information criteria, yielding an average inclusion of less than one irrelevant variable across almost all tested scenarios. Conversely, the framework inclusively captured over 99\% of the relevant variables in all tested scenarios.

Moreover, the benchmark models faltered in maintaining performance efficacy as the numbers of candidates and true features escalated. This disparity underscores the superior adaptability and robustness of the State Space Learning framework compared to traditional benchmarks, especially under conditions of increased complexity and dimensionality.

Table \ref{TableSim2} presents the results of the average of the Monte Carlo simulation for MSE (average mean squared error), Bias and execution time for each model. The values of MSE and Bias are calculated as:

\begin{align}
    MSE = \frac{1}{500\cdot p}\sum_{i=1}^p\Biggl(\sum_{j=1}^{500}(\hat{\beta}_{i, j} - \beta_{i, j})^2\Biggl)\\
    Bias = \frac{1}{500\cdot p}\sum_{i=1}^p\Biggl(\sum_{j=1}^{500}(\hat{\beta}_{i, j} - \beta_{i, j})\Biggl)
\end{align}

\begin{table}[H]
\footnotesize
\centering
\begin{tabular}{|c|cc|cc|cc|cc|cc|cc|}
\hline
\multicolumn{1}{c}{} & \multicolumn{2}{c}{SSL (AIC)} & \multicolumn{2}{c}{SSL (BIC)} & \multicolumn{2}{c}{SS+AdaLasso (AIC)} & \multicolumn{2}{c}{SS+AdaLasso (BIC)} & \multicolumn{2}{c}{SS+FW (AIC)} & \multicolumn{2}{c}{SS+FW (BIC)} \\ \hline
\multicolumn{13}{c}{}                                                                      \\
M\textbackslash{}K   & 50                      & 100                  &                       50                      & 100                  &                       50                 & 100            &                       50                 & 100    
&                       50                 & 100  
&                       50                 & 100  
\\ \hline
\multicolumn{13}{|c|}{Panel (a): Average MSE}                     \\ \hline
3 & 2.4e-4 & 2.0e-4 & 2.4e-4 & 2.0e-4 & 0.046 & 1.3e7 & 0.013 & 84.272 & 0.024 & 2.7e7 & 0.024 & 2.7e7 \\ 
5 & 0.001 & 6.1e-4 & 0.001 & 6.1e-4 & 3.2e2 & 97.568 & 0.03 & 0.028 & 0.058 & 0.034 & 0.058 & 0.034 \\ 
8 & 0.002 & 0.001 & 0.002 & 0.001 & 0.131 & 0.049 & 0.066 & 0.049 & 0.111 & 0.059 & 0.113 & 0.06 \\ 
10 & 0.003 & 0.004 & 0.003 & 0.004 & 4.3e4 & 0.072 & 0.09 & 0.071 & 2.9e5 & 0.083 & 0.157 & 0.085 \\ \hline
\multicolumn{13}{|c|}{Panel (b): Average Bias}                                                                          \\ \hline
3 & -7.2e-4 & -4.3e-4 & -7.7e-4 & -4.5e-4 & -2.0e-2 & 34.635 & -2.2e-2 & 0.029 & -2.4e-2 & 23.456 & -2.4e-2 & 23.456 \\ 
5 & -1.9e-3 & -1.1e-3 & -2.0e-3 & -1.2e-3 & 0.115 & 0.013 & -4.0e-2 & -2.1e-2 & -5.4e-2 & -2.7e-2 & -5.4e-2 & -2.7e-2 \\ 
8 & -4.5e-3 & -2.1e-3 & -4.6e-3 & -2.1e-3 & -5.9e-2 & -3.7e-2 & -6.8e-2 & -3.7e-2 & -9.9e-2 & -5.0e-2 & -10.0e-2 & -5.1e-2 \\ 
10 & -6.3e-3 & -6.1e-3 & -6.5e-3 & -6.1e-3 & -1.0e-1 & -5.1e-2 & -8.5e-2 & -5.2e-2 & -3.5e0 & -6.8e-2 & -1.3e-1 & -6.9e-2 \\  \hline
\multicolumn{13}{|c|}{Panel (c): Median MSE}                     \\ \hline
3 & 2.6e-5 & 1.2e-5 & 2.8e-5 & 1.3e-5 & 0.039 & 0.016 & 0.009 & 0.015 & 0.02 & 0.01 & 0.02 & 0.01  \\ 
5 & 5.6e-5 & 2.7e-5 & 5.7e-5 & 2.8e-5 & 0.068 & 0.027 & 0.022 & 0.027 & 0.06 & 0.03 & 0.06 & 0.03 \\ 
8 & 1.4e-4 & 6.4e-5 & 1.4e-4 & 6.8e-5 & 0.114 & 0.047 & 0.051 & 0.047 & 0.108 & 0.06 & 0.112 & 0.06 \\ 
10 & 2.3e-4 & 1.2e-4 & 2.4e-4 & 1.2e-4 & 0.134 & 0.067 & 0.072 & 0.067 & 0.152 & 0.082 & 0.161 & 0.083 \\
  \hline
\multicolumn{13}{|c|}{Panel (d): Median MSE}                     \\ \hline
3 & -4.2e-4 & -1.8e-4 & -4.4e-4 & -1.8e-4 & -2.1e-2 & -1.2e-2 & -2.0e-2 & -1.1e-2 & -2.0e-2 & -1.0e-2 & -2.0e-2 & -1.0e-2 \\ 
5 & -8.3e-4 & -4.4e-4 & -8.5e-4 & -4.7e-4 & -3.8e-2 & -2.0e-2 & -3.8e-2 & -2.0e-2 & -5.7e-2 & -2.8e-2 & -5.7e-2 & -2.8e-2 \\ 
8 & -2.0e-3 & -1.0e-3 & -2.0e-3 & -1.1e-3 & -5.7e-2 & -3.6e-2 & -6.2e-2 & -3.6e-2 & -9.9e-2 & -5.0e-2 & -9.9e-2 & -5.1e-2 \\ 
10 & -3.3e-3 & -1.9e-3 & -3.4e-3 & -1.9e-3 & -6.5e-2 & -5.0e-2 & -7.7e-2 & -5.0e-2 & -1.3e-1 & -6.9e-2 & -1.3e-1 & -7.0e-2 \\
 \hline
\multicolumn{13}{|c|}{Panel (e): Average Time}     
\\ \hline
3 & 0.013 & 0.015 & 0.013 & 0.014 & 0.146 & 0.15 & 0.145 & 0.151 & 4.9e1 & 1.0e2 & 4.7e1 & 9.6e1 \\ 
5 &   0.014 & 0.016 & 0.012 & 0.014 & 0.154 & 0.153 & 0.153 & 0.152 & 5.6e1 & 1.1e2 & 5.4e1 & 1.1e2 \\ 
8 &  0.013 & 0.017 & 0.013 & 0.015 & 0.157 & 0.148 & 0.157 & 0.149 & 6.6e1 & 1.3e2 & 6.1e1 & 1.2e2 \\ 
10 &  0.013 & 0.015 & 0.013 & 0.016 & 0.151 & 0.151 & 0.151 & 0.151 & 6.6e1 & 1.3e2 & 5.7e1 & 1.1e2  \\  \hline
    \end{tabular}
    \caption{Statistics concerning model estimation processes. Panel (a), Average MSE estimated. Panel (b) Average Bias estimated. Panel (c), Median MSE estimated. Panel (f) Median Bias estimated. (e) Average run time.}
\label{TableSim2}
\end{table}

The proposed framework significantly outperformed the benchmarks in terms of average Mean Squared Error (MSE) and bias across all tested configurations. This substantial improvement suggests that the estimated models of the SSL more accurately reflect the true coefficients of the exogenous features.

Notably, a minority of the test samples produced exceptionally poor results in terms of bias and MSE for the benchmarks. However, even when assessing the median values of these metrics, the proposed framework consistently demonstrated superior performance. This finding indicates that the SSL framework exhibits greater robustness to challenging samples and also for the average scenarios.

Moreover, It is imperative to underscore the substantial enhancement in computational efficiency demonstrated by the proposed framework, surpassing the forward benchmark by over 1000 times in all tested scenarios and surpassing the stepwise Adaptive Lasso by over 5 times in all tested scenarios. This marked high computation cost of the forward selection method can be attributed to the inherent computational demands, necessitating the estimation of a considerable number of models contingent upon the volume of candidate variables.

\section{Conclusion}

This work introduced State Space Learning (SSL), a comprehensive framework designed to address some relevant limitations of traditional state-space models in time series analysis. More specifically, through a novel formulation that incorporates high-dimensional regularized regression, the proposed SSL approach not only improves the extraction of unobserved components but also efficiently manages outlier detection and optimal subset selection, even in non-stationary time series. Additionally, the regularized regression format allows for a polynomial-time estimation process with global optimality guarantees. Therefore, SSL significantly enhances the flexibility and robustness of time series modeling, particularly in high-dimensional settings, providing a better representation of systems through state-space models that effectively tackle modern data challenges.

The numerical experiments, conducted on both synthetic and real-world datasets, demonstrate the predictive power and robustness of the SSL framework. Based on the results obtained using the M4 competition dataset, the superiority of our approach over traditional benchmarks (such as Kalman filtering and ARIMA models) in both forecasting and subset selection activities corroborates the robustness of our contribution. 

By extending state-space modeling to a framework that can be applied to any linear state-space model with time-varying coefficients, SSL paves the way for broader applicability in fields beyond time series analysis. The implementation of our methodology in the open-source package ``StateSpaceLearning.jl'' ensures that researchers and practitioners alike can access and apply this framework to diverse temporal data problems.

Based on the provided evidence, the proposed SSL framework represents a significant step forward in the evolution of time series modeling, bridging the gap between classical state-space models and modern machine learning techniques. Future research could explore further extensions to non-linear dynamics and scaling improvements, continuing to refine the framework and expand its scope of applications.

\bibliographystyle{plainnat}
\bibliography{bibtex}

\end{document}